\documentclass[runningheads]{llncs}
\usepackage[dvipsnames]{xcolor}
\usepackage[T1]{fontenc}
\usepackage{graphicx}
\usepackage{tikz}
\usepackage{comment}
\usepackage{amsmath,amssymb}
\usepackage{color}
\usepackage{nicefrac}
\usepackage{enumitem}

\usepackage{booktabs}
\usepackage{multirow}
\usepackage{xspace}
\usepackage{rotating}

\usepackage[bottom]{footmisc}
\usepackage[pagebackref,breaklinks,colorlinks,citecolor=ForestGreen]{hyperref}

\usepackage{soul}
\usepackage{wrapfig}

\newcommand{\myparagraph}[1]{\vspace{0pt}\noindent\textbf{#1}\xspace}

\newcommand{\alert}[1]{{\color{red}{#1}}}

\newcommand{\ie}{\textit{i.e.}\xspace}
\newcommand{\etal}{\textit{et al.}\xspace}

\newcommand{\ssvtwo}{SS-v2\xspace} 
\newcommand{\kinetics}{Kinetics-100\xspace} 
\newcommand{\ucf}{UCF-101\xspace} 

\newcommand{\oneshot}{1-shot\xspace} 
\newcommand{\fiveshot}{5-shot\xspace}

\newcommand{\stddev}[1]{\scriptsize{$\pm#1$}}

\newcommand{\diffup}[1]{{\color{OliveGreen}{($\uparrow$ #1)}}}
\newcommand{\diffdown}[1]{{\color{BrickRed}{($\downarrow$ #1)}}}

\def\l1{\ensuremath{\ell_1}\xspace}
\def\l2{\ell_2\xspace}

\newcommand{\real}{\mathbb{R}}

\def\nsp{\hspace{-3pt}}
\def\xssp{\hspace{1pt}}
\def\ssp{\hspace{3pt}}
\def\msp{\hspace{6pt}}

\newcommand{\vq}{\mathbf{q}}

\newcommand{\vt}{\mathbf{t}}

\newcommand{\vx}{\mathbf{x}}

\makeatletter
\def\wrapfill{\par
\ifx\parshape\WF@fudgeparshape
\nobreak
\ifnum\c@WF@wrappedlines>\@ne
\advance\c@WF@wrappedlines\m@ne
\vskip\c@WF@wrappedlines\baselineskip
\global\c@WF@wrappedlines\z@
\fi
\allowbreak
\WF@finale
\fi
}
\makeatother
\usepackage{pgfplots, pgfplotstable}
\pgfplotsset{compat=newest}
\usepgfplotslibrary{fillbetween}

\newcommand{\resnetc}{NavyBlue}
\newcommand{\rtwoplusonec}{Orange}

\newcommand{\resnetfvc}{Gray}
\newcommand{\rtwoplusonefvc}{Aquamarine}

\newcommand{\markertemporal}{triangle*}
\newcommand{\markernontemporal}{*}
\newcommand{\markerimplicitlytemporal}{\markertemporal}
\newcommand{\markerspatiotemporal}{diamond*}
\newcommand{\markerparametric}{o}
\newcommand{\markerobject}{square*}

\pgfmathsetmacro{\teasermarkersize}{3.5}
\pgfmathsetmacro{\stdgrad}{10}

\tikzset{every mark/.append style={solid}}
\pgfplotsset{
	grid=both, width=\linewidth, try min ticks=5,
	legend cell align=left, legend style={fill opacity=0.8},
	ylabel near ticks,
    xlabel near ticks,
    every tick label/.append style={font=\footnotesize},
}

\pgfplotsset{
    resnet/.style={thick, color=\resnetc, mark=square*,mark size=\teasermarkersize pt, only marks},
    rtwoplusone/.style={thick, color=\rtwoplusonec, mark=square*,mark size=\teasermarkersize pt, only marks},
    cmn/.style={thick, color=\resnetc, mark=\markernontemporal,mark size=\teasermarkersize pt, only marks},
    otamresnet/.style={thick, color=\resnetc, mark=\markertemporal,mark size=\teasermarkersize pt, only marks},
    pal/.style={thick, color=\resnetc, mark=\markertemporal,mark size=\teasermarkersize pt, only marks},
    titan/.style={thick, color=\resnetc, mark=\markertemporal,mark size=\teasermarkersize pt, only marks},
    trxresnet/.style={thick, color=\resnetc, mark=\markerimplicitlytemporal,mark size=\teasermarkersize pt, only marks},
    mtfan/.style={thick, color=\resnetc, mark=\markerspatiotemporal,mark size=\teasermarkersize pt, only marks},
    hyrsm/.style={thick, color=\resnetc, mark=\markerspatiotemporal,mark size=\teasermarkersize pt, only marks},
    cpm/.style={thick, color=\resnetc, mark=\markerobject,mark size=\teasermarkersize pt, only marks},
    tap/.style={thick, color=\resnetc, mark=\markertemporal,mark size=\teasermarkersize pt, only marks},
    bmvc/.style={thick, color=\resnetc, mark=\markerparametric,mark size=\teasermarkersize pt, only marks},
    tsl/.style={thick, color=\rtwoplusonec, mark=\markerparametric,mark size=\teasermarkersize pt, only marks},
    trx/.style={thick, color=\rtwoplusonec, mark=\markerimplicitlytemporal,mark size=\teasermarkersize pt, only marks},
    mean/.style={thick, color=\rtwoplusonec, mark=\markernontemporal,mark size=\teasermarkersize pt, only marks},
    max/.style={thick, color=\rtwoplusonec, mark=\markernontemporal,mark size=\teasermarkersize pt, only marks},
    diag/.style={thick, color=\rtwoplusonec, mark=\markertemporal,mark size=\teasermarkersize pt, only marks},
    linear/.style={thick, color=\rtwoplusonec, mark=\markertemporal,mark size=\teasermarkersize pt, only marks},
    chamfer/.style={thick, color=\rtwoplusonec, mark=\markernontemporal,mark size=\teasermarkersize pt, only marks},
    otam/.style={thick, color=\rtwoplusonec, mark=\markertemporal,mark size=\teasermarkersize pt, only marks},
    visil/.style={thick, color=\rtwoplusonec, mark=\markertemporal,mark size=\teasermarkersize pt, only marks},
}

\pgfplotsset{
    resnetfv/.style={thick, color=\resnetfvc, mark=square*,mark size=\teasermarkersize pt, only marks},
    rtwoplusonefv/.style={thick, color=\rtwoplusonefvc, mark=square*,mark size=\teasermarkersize pt, only marks},
    cmnfv/.style={thick, color=\resnetfvc, mark=\markernontemporal,mark size=\teasermarkersize pt, only marks},
    otamresnetfv/.style={thick, color=\resnetfvc, mark=\markertemporal,mark size=\teasermarkersize pt, only marks},
    trxresnetfv/.style={thick, color=\resnetfvc, mark=\markerimplicitlytemporal,mark size=\teasermarkersize pt, only marks},
    bmvcfv/.style={thick, color=\resnetfvc, mark=\markerparametric,mark size=\teasermarkersize pt, only marks},
    tslfv/.style={thick, color=\rtwoplusonefvc, mark=\markerparametric,mark size=\teasermarkersize pt, only marks},
    trxfv/.style={thick, color=\rtwoplusonefvc, mark=\markerimplicitlytemporal,mark size=\teasermarkersize pt, only marks},
    meanfv/.style={thick, color=\rtwoplusonefvc, mark=\markernontemporal,mark size=\teasermarkersize pt, only marks},
    diagfv/.style={thick, color=\rtwoplusonefvc, mark=\markertemporal,mark size=\teasermarkersize pt, only marks},
    otamfv/.style={thick, color=\rtwoplusonefvc, mark=\markertemporal,mark size=\teasermarkersize pt, only marks},
    visilfv/.style={thick, color=\rtwoplusonefvc, mark=\markertemporal,mark size=\teasermarkersize pt, only marks},
    tstfv/.style={thick, color=\rtwoplusonefvc, mark=\markerimplicitlytemporal,mark size=\teasermarkersize pt, only marks},
}

\usepackage{subcaption}

\begin{document}
\title{Rethinking matching-based \\ few-shot action recognition}
\titlerunning{Rethinking matching-based few-shot action recognition}
\authorrunning{J. Bertrand et al.}

\author{Juliette Bertrand\inst{1}
\and
Yannis Kalantidis\inst{2} 
\and
Giorgos Tolias\inst{1}
}

\authorrunning{Bertrand et al.}
\institute{Visual Recognition Group, Faculty of Electrical Engineering, \\ Czech Technical University in Prague \and
NAVER LABS Europe, Grenoble, France}

\maketitle
\begin{abstract}
Few-shot action recognition, \ie recognizing new action clas\-ses given only a few examples, benefits from incorporating temporal information. Prior work either encodes such information in the representation itself and learns classifiers
at test time, or obtains frame-level features and performs pairwise temporal matching.
We first evaluate a number of matching-based approaches using features from spatio-temporal backbones, a comparison missing from the literature, and show that the gap in performance between simple baselines and more complicated methods is significantly reduced. Inspired by this, we propose Chamfer++, a non-temporal matching function that achieves state-of-the-art results in few-shot action recognition. We show that, when starting from temporal features, our parameter-free and interpretable approach can outperform all other matching-based and classifier methods for one-shot action recognition on three common datasets without using temporal information in the matching stage.

\vspace{1pt}
Project page: \url{https://jbertrand89.github.io/matching-based-fsar}

\end{abstract}
\section{Introduction}
\label{sec:introduction}

Recognizing actions within videos is essential for analyzing trends, enhancing broadcasting experience, or filtering out inappropriate content. 
However, collecting and annotating enough video examples to train supervised models can be prohibitively time-consuming. 
It is therefore desirable to recognize new action classes with as few labeled examples as possible. 
This is the premise behind the task of few-shot learning, where models learn to adapt to a set of unseen classes for which only a few examples are available. 
In video action recognition, additional challenges arise from the temporal dimension. 
Recognition methods need to capture the scene's temporal context and temporal dynamics.

\begin{figure}
  \vspace{-20pt}  
\centering
  \begin{subfigure}{0.75\textwidth}
    \includegraphics[width=\textwidth]{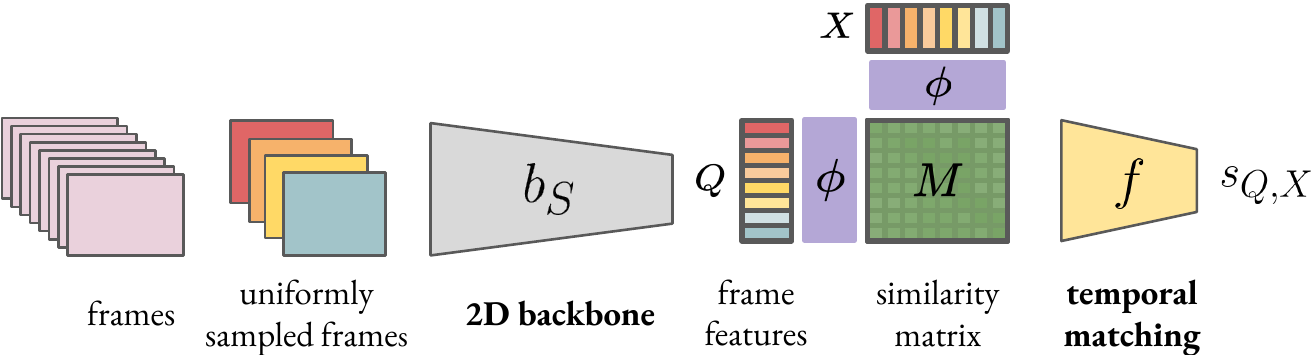}
     \vspace{-15pt}
    \caption{}
    \label{fig:first}
\end{subfigure}     
\vspace{2pt}
\begin{subfigure}{0.75\textwidth}
\vspace{-0pt}
    \includegraphics[width=\textwidth]{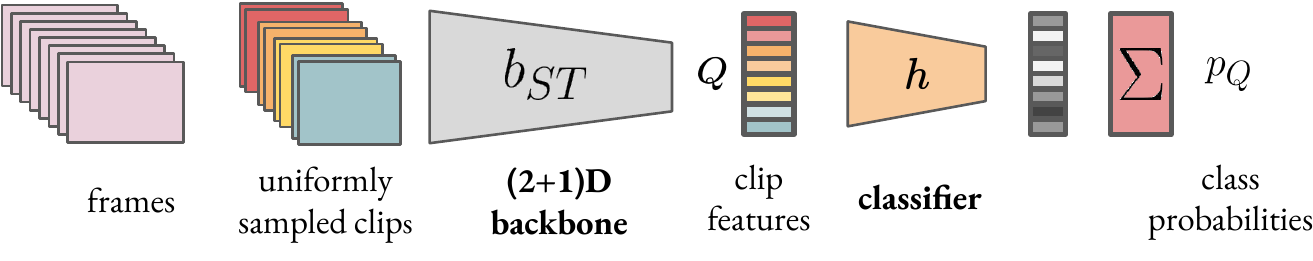}
    \vspace{-20pt}
    \caption{}
    \label{fig:second}
\end{subfigure}
\begin{subfigure}{0.75\textwidth}
\vspace{-8pt}
    \includegraphics[width=\textwidth]{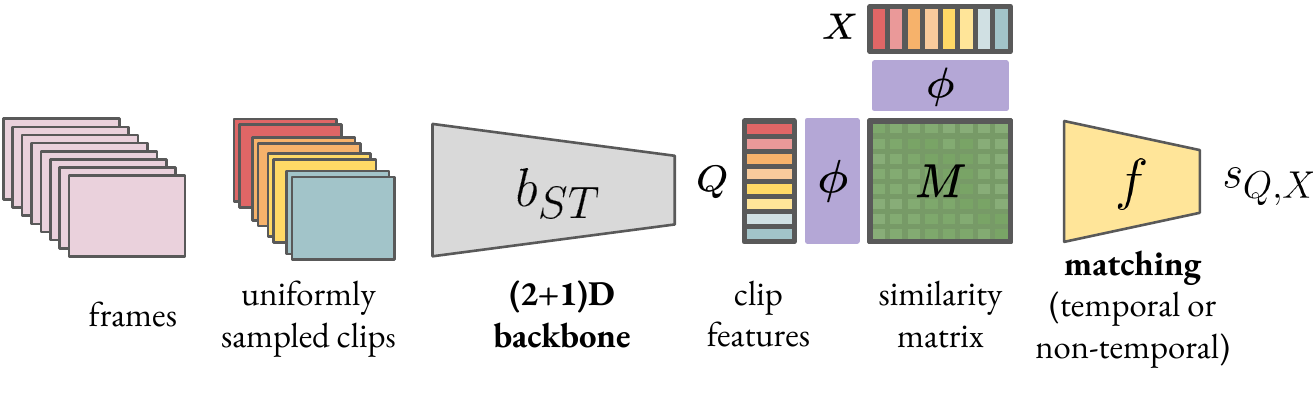}
    \vspace{-22pt}
    \caption{}
    \label{fig:third}
\end{subfigure}
  \vspace{-5pt}
\caption{\textbf{Different ways of using temporal information in few-shot action recognition.} Temporal information is utilized: a) during matching (existing \textit{matching-based} methods) b) in the backbone with a classifier (existing \textit{classifier-based} methods, \emph{e.g.}~\cite{tsl}) c) in the backbone with a matching step (this paper). 
  \label{fig:spatio_temporal}
\vspace{-15pt}
}
\end{figure}

One family of approaches is formed by \emph{matching-based} methods \cite{otam,trx,Zhu2021pal,thatipelli2021spatio}
where each
test example or ``query'' is compared against all support examples of a class to infer a class confidence score. 
Most existing matching-based methods use frame-level representations, \ie a 2D convolutional backbone that takes a frame as input, and a feature set is formed by encoding multiple frames. Feature extraction is followed by matching the query feature set $Q$ to the support example set $X$, and a similarity $s_{Q,X}$ between the two is computed.
This process is depicted in Figure~\ref{fig:first}. 
Although each feature represents an individual frame and cannot capture temporal information, the feature sets are usually temporal sequences, and the matching process can exploit such information.

Another family of approaches is formed by methods that learn a conventional linear \emph{classifier} at test time, \ie using the handful of examples available. In this case, any temporal context has to be incorporated in the representation, as shown in Figure~\ref{fig:second}.
As a representative example, 
Xian~\etal~\cite{tsl} adopt the spatio-temporal R(2+1)D architecture~\cite{r2plus1d}, where
the input is a video \textit{clip}, \ie a sequence of consecutive frames, and convolutions across the temporal dimension enable the features to encode temporal information.
Following findings in few-shot learning~\cite{wang2019simpleshot,chen2020closer}, Xian~\etal further abandon episodic training and instead fine-tune a pre-trained backbone using all training examples of the base classes. 
Using strong temporal features and by simply learning a linear classifier at test time, they report state-of-the-art results for few-shot action recognition.

Motivated by the two families of approaches presented above, we introduce a new setup depicted in Figure~\ref{fig:third} that aims at answering the following questions:

\noindent\emph{1. Do matching-based methods still have something to offer for few-shot action recognition given strong temporal representations?} 
We level the playing field with respect to representations and evaluate a number of recent matching-based approaches using strong temporal representations. We find that such approaches perform better than training a classifier at test time.

\noindent\emph{2. Is temporal matching necessary when the features capture temporal information?} We show that matching-based methods invariant to the temporal order in the feature sequence (\emph{non-temporal matching}) are performing as good as the ones that do use it (\emph{temporal matching}) on many common benchmarks. 

Inspired by the findings above, we further introduce \textbf{Chamfer++}, a novel, parameter-free and interpretable matching approach that employs Chamfer matching and is able to achieve a new state-of-the-art for one-shot action recognition on three common benchmarks.

\section{Related Work}
\label{sec:related}

In the image domain, several methods for few-shot classification are metric-learning-based~\cite{vinyals2016matching,Snell2017Proto,Sung2018Compare,Doersh2020CrossTransformers,cao21comet} and use a k-nearest-neighbor classifier at test time. 
Despite the dominance of meta-learning-based methods in the area~\cite{finn2017model,Finn2018,Grant2018,Rusu2019LatentEmbedding,lifchitz2019dense}, several recent studies highlight the importance of starting from strong visual representations. It is shown~\cite{wang2019simpleshot,chen2020closer} that if one learns representations without meta-learning but using instead all available data from all base classes, simple nearest-neighbors~\cite{wang2019simpleshot} or parametric classifier~\cite{chen2020closer} baselines work on-par or better than most methods on common benchmarks. Similar observations are made for few-shot learning in the video domain~\cite{tsl,Zhu2021CloserLook,Zhu2021pal}.

As with the image domain, most methods leverage meta-learning and are grouped into initialization-based~\cite{tsl,Zhu2021CloserLook},  metric-learning~\cite{zhu2018cmn,bishay2019tarn,otam,Zhang2020FewShotAR,trx,Zhu2021pal} and generative-based~\cite{Dwivedi2019Protogan,tsl} methods. 
However, unlike the image domain, most methods for few-shot action recognition try to take into account the \textit{temporal} dimension in the visual representations and/or during the matching process. In fact, most methods incorporate temporal matching~\cite{zhu2018cmn,otam,bishay2019tarn,visil,Zhang2020FewShotAR,trx,li2021ta2n}.
The features are either extracted from single frames~\cite{zhu2018cmn,otam,trx} or clips~\cite{bishay2019tarn,tsl,Zhang2020FewShotAR} where temporal information is already captured in the features.

Some matching methods explicitly aim for temporal alignment and estimate video-to-video similarity via an ordered temporal alignment score.
OTAM~\cite{otam} finds the optimal path on the temporal similarity matrix via a differentiable approximation of the Dynamic Time Warping (DTW)~\cite{Muller2007DTW} algorithm, a method also used for alignment in other temporal tasks~\cite{chang2019d3tw,dvornik2021drop}. 
Other methods like ARN~\cite{Zhang2020FewShotAR} use spatio-temporal attention, while TA$^2$N~\cite{li2021ta2n} proposes two-stage spatial-temporal alignment.
TARN~\cite{bishay2019tarn} uses temporal attention over clip sequences for alignment. During the training process, the attention parameters, together with the parameters from a subsequent recurrent network, are learned such that features are aligned between the query and a support video from the correct class.
 
Other methods like TRX \cite{trx} or PAL \cite{Zhu2021pal} do not explicitly seek alignment but rely on \textit{cross-attention} mechanisms over the query and support features to perform temporal matching. In both cases, class-wise representations are constructed and matching is performed directly in a \textit{video-to-class} manner. For TRX, class representations are adapted on-the-fly in a query-conditioned manner, while for PAL the query features are instead adapted to match pre-computed class-wise representations. 
In our study, we show how such direct video-to-class approaches are highly competitive when more than one videos per class are provided, but that video-to-video methods are superior and more efficient for the one-shot case. 

Recently, a number of methods suggest that the meta-learning framework is not optimal for the representation learning phase. Following similar observations in the image classification domain~\cite{wang2019simpleshot,chen2020closer}, some few-shot action recognition methods~\cite{tsl,Zhu2021CloserLook,Zhu2021pal} learn their representation on the full train set and report higher performance on multiple benchmarks. As discussed in Section~\ref{sec:framework}, we follow such methods, and use a strong representation learned on the entire train set as the starting point for our study on temporal matching.
\section{Method}
\begin{figure*}[t]
\vspace{-5pt}

\begin{subfigure}[t]{0.27\textwidth}
    \centering
    \includegraphics[width=.9\linewidth]{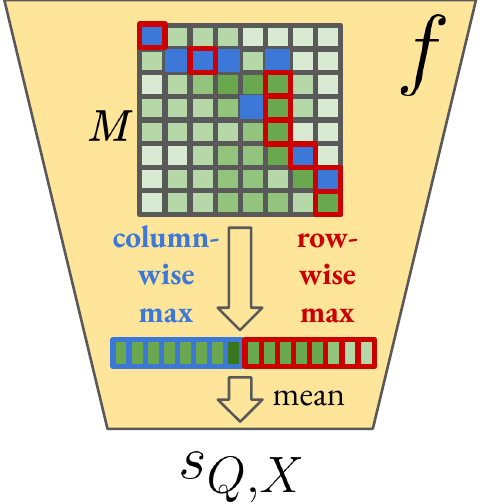}
    \caption{Chamfer matching}
    \label{fig:chamfer_matching}
\end{subfigure}
\begin{subfigure}[t]{0.24\textwidth}
    \centering
    \includegraphics[width=.85\linewidth]{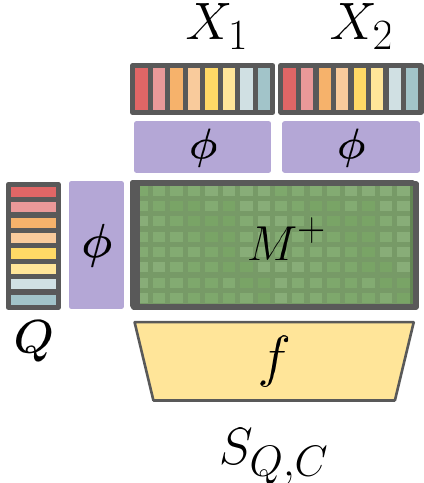}
    \caption{Joint matching}
    \label{fig:joint_matching}
\end{subfigure}
\hspace{10pt}
\begin{subfigure}[t]{0.45\textwidth}
    \centering
    \includegraphics[width=\linewidth]{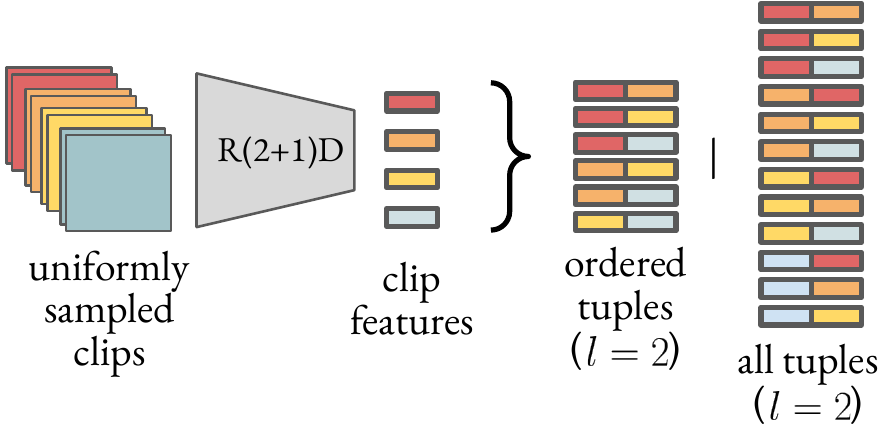}
    \caption{Clip tuples}
    \label{fig:clip_tuples}
\end{subfigure}  

\vspace{3ex}
    \caption{\textbf{Details of the proposed matching approach}. a) The \textbf{Chamfer} matching function $f_{QS}$ from Eq.(\ref{eq:chamferqs}). b) Jointly matching multiple examples per class (Chamfer+). c) using  clip tuples as representation; one can use only ordered tuples or all tuples; in both cases, the matching part remains \textit{non-temporal}, \ie invariant to the temporal order of features.
    \vspace{-3ex}
    }
    \label{fig:method}
\end{figure*}

We first present the video representation and the experimental protocol used in this work in Section~\ref{sec:preliminaries}. Then, we describe a framework to fairly compare matching-based and classifier-based approaches in Section~\ref{sec:framework} and propose a new matching function in Section~\ref{sec:chamferqs}.
In the following, we refer to videos as examples and actions as classes.
\label{sec:method}


\subsection{Preliminaries}
\label{sec:preliminaries}
We describe the video representation that we use and the episodic protocol as formulated in the recent literature of few-shot video action recognition, which we follow in this work.

\subsubsection{Video representation}
\label{sec:videorepresentation}

A clip $c_i$ is a sequence of $L$ consecutive RGB frames in the form of a $L \times H \times W \times 3$ tensor. A deep video backbone $b$ takes a clip $c_i$ as input and maps it to a $d$-dimensional vector $\vq_i = b(c_i) \in \real^d$, named feature.
We use the R(2+1)D backbone architecture~\cite{r2plus1d} as in the work of Xian \etal~\cite{tsl}, which uses efficient and effective separated spatio-temporal convolutions.
A video $Q$ is represented by a set of clip features $Q=\{\vq_i\}$. The clips are uniformly sampled over the temporal dimension, with possible overlap.

\subsubsection{Episodic protocol}
\label{sec:episodic}

We adopt the commonly used setup~\cite{trx,otam}. 
The classes of the train and test sets are non-overlapping and usually named base and novel classes, respectively. 
To simulate the limited annotated data, test episodes are randomly sampled from the test set. Each episode corresponds to a different classification task and comprises query and support examples for a fixed set of classes, where labels of support examples are known, while labels of query examples are unknown. Only $k$ labeled examples per class, also named shots, are available in the support set, with $k$ typically ranging from 1 to 5. The performance is evaluated via classification accuracy on the query examples averaged over all test episodes.

\subsection{A common setup for classifier and matching-based approaches}
We identify two dominant families of approaches proposed in the recent few-shot action recognition literature: the classifier-based and the matching-based methods \footnote{Prototypical networks can be seen as an extension of matching based methods \cite{trx,Zhu2021pal,thatipelli2021spatio}. Hence we group them with the matching-based family.}. Unfortunately, discrepancies in setup and architecture between existing approaches make it difficult to compare them fairly. 
We propose to follow the same representation learning strategy and start from a common frozen backbone.

\subsubsection{Classifier-based approaches}
\label{sec:classifier}
The classifier-based approaches ~\cite{tsl,Zhu2021CloserLook} train the video representation using a classifier in the form of a linear layer. This is similar to the corresponding work on few-shot learning in the image domain~\cite{GK18}.

Classifiers are trained using the whole dataset, but episodic training cannot guarantee to see all the examples because it is intractable to sample all the possible episodes.
Xian \etal~\cite{tsl} depart from episodic training and propose a \textit{Two-stage Learning} (\textit{TSL}) process. 
During the first stage, a R(2+1)D video backbone and a classifier are learned jointly using all the labeled examples of the train set. During the second stage, the backbone remains fixed to avoid overfitting and  a newly initialized classifier needs to be trained per test episode using  the support examples. 

In both stages, a linear classifier with a soft-max function denoted by $h: \real^d \rightarrow \real^C$, where C is the number of classes, is added to the output of the backbone. During the training stage, $C=C_{t}$ is equal to the number of all classes in the train set. During the second stage, $C=C_{f}$ is equal to the number of classes per test episode, usually $C_{f}=5$.
Training is performed by optimizing the class probabilities $h(\vq_i)$ for each $\vq_i \in Q$ with the cross-entropy loss ($\mathcal{L}_{cls}$), while inference is performed by sum-pooling of the classifier output across clips, \ie $\sum_{\vq_i \in X} h(\vq_i)$.

\subsubsection{Matching-based approaches}
\label{sec:matching}
The matching-based approaches estimate the similarity between the query and all the support examples of each class to obtain class probabilities. 
Training is performed on episodes sampled from the train set, which are meant to imitate the episodes of the test set.

Let $Q=\{\vq_i\}$ and $X=\{\vx_i\}$ be two videos with $|Q|=|X|=n$ and assume that $n$ is constant across videos.
We form the \emph{temporal similarity matrix} for the ordered video pair $(Q,X)$ denoted by $M_{Q,X}\in \real^{n\times n}$, or just $M$ for brevity, with elements $m_{ij} = \phi(\vq_i)^\top \phi(\vx_j)$, where $\phi: \real^d \rightarrow \real^D$ is a learnable projection head.
The function $\phi$ consists of a linear layer, a layer normalization, and a $\ensuremath{\ell_2}$  normalization to guarantee bounded similarity values $m_{ij}$. When no linear projection is used, $\phi$ is equivalent to the identity mapping with an \ensuremath{\ell_2} normalization.
Each element $m_{ij}$ of the matrix $M$ can be seen as a temporal correspondence between clip $i$ of video $Q$ and clip $j$ of video $X$. 

We consider the family of matching approaches that infer a video-to-video similarity $S_{Q, X}$, between video $Q$ and $X$, solely based on the matrix $M$, \ie $S_{Q,X} = f(M)$, with $f:\real^{n \times n} \rightarrow \real$, named the \emph{matching function}.
The scalar $S_{Q,X}$ should be high if the two videos depict the same action. 
By definition, the result of function $f$ only depends on the strength and position of the pairwise similarities $m_{ij}$ and does not directly depend on the features themselves.
The function $f$ can either be hand-crafted or include learnable parts.
A graphical overview of different matching approaches is given in the 
appendix as long as a more detailed description of them.
Some matching functions use temporal information by leveraging the position, either absolute or relative, of the pairwise similarities $m_{ij}$. The matching functions that use temporal information are called \emph{temporal}, whereas the others are called \emph{non temporal}.

The pairwise video-to-video similarities between query and support examples are averaged per class to obtain class probabilities
\footnote{In prototypical networks  \cite{trx,Zhu2021pal,thatipelli2021spatio}, the pairwise clip-to-clip similarities are used as weights to compute a class prototype specific to the query example. The class probabilities are computed as the distance between a query and its prototype.}.
During training, the class probabilities are optimized with the cross-entropy loss ($\mathcal{L}_{cls}$). Inference is also performed by estimating the similarity between query and all support examples. This is a form of a k-nearest-neighbor classifier.

\subsubsection{A common starting point}
\label{sec:framework}

We follow the training stage of TSL~\cite{tsl} to learn the R(2+1)D backbone parameters using all the annotated examples. We freeze this backbone, and treat the resulting model as a feature extractor. This is our starting point for both classifier-based and matching-based approaches which enable us to fairly compare the two families of approaches. 
Specifically, matching-based methods only learn the feature projection function $\phi$ and the matching parameters when needed in a test-agnostic way. Unlike classifier-based methods, which need to train a classifier for every testing episode, matching-based approaches require no learning or adaptation at test time. They only need the pairwise matching between the query and each one of the support examples.

\subsection{Chamfer++}
\label{sec:newchamfer}
We propose a new matching function, Chamfer++, which is non-temporal and achieves top performance while being parameter-free and intuitive. 
It is an extension of Chamfer similarity (Figure~\ref{fig:chamfer_matching}) with joint-matching over multiple shots  (Figure~\ref{fig:joint_matching}) applied in conjunction with clip-feature-tuples instead of clip features (Figure~\ref{fig:clip_tuples}). This section details its main components.

\subsubsection{Chamfer}
\label{sec:chamferqs}
The Chamfer matching function $f_{Q}$ is given by
\begin{equation}
f_{Q}(M) := \frac{1}{n}\sum_i \max_j m_{ij}. 
\end{equation}
Each clip of the query example is matched with its most similar one within the clips of a support example to produce the score between this specific query clip and this support example. The final video-to-video similarity is the average of all the query clip scores.
The Chamfer matching function implies that each clip sampled from the query example contributes to the similarity score by matching its closest clip in the support example. 
One can transpose the temporal similarity matrix and derive the symmetric process where each clip from the support example needs to match a query clip. Then the matching function becomes
\begin{equation}
f_{S}(M) := \frac{1}{n}\sum_j \max_i m_{ij}.
\end{equation}
Chamfer-S is equivalent to the Chamfer on the transposed matrix $M^\top$. 
Summing the two gives a symmetric Chamfer variant, where all clips from both the query and the support example are required to match: 
\begin{equation}
f_{QS}(M) := f_{Q}(M) + f_{S}(M).
\label{eq:chamferqs}
\end{equation}
We refer to this symmetric variant as simply \emph{Chamfer matching} in the context of few-shot action recognition.
In the following, we also refer to Chamfer-Q as query-based and Chamfer-S as support-based.

\subsubsection{Joint-matching}
\label{sec:jointmatching}

As discussed in Section~\ref{sec:matching}, the standard option to compute the query-to-class probabilities is by averaging the pairwise similarity score between the query example and all the support examples belonging to the class.
Instead, we propose to match all support examples jointly.
We concatenate the temporal similarity matrices between the query and all support examples. It creates the joint temporal similarity matrix 
$M^{+}$, with $M^{+}\in \real^{n \times kn}$.
Then, we compute the matching function on top of 
$M^{+}$
to obtain video-to-class similarity 
$S_{Q,c} := s(M^{+})$. 
We evaluate the impact of this newly proposed joint-matching versus the standard  single-matching in Section~\ref{sec:ablation}. We refer to this variant as Chamfer+ in the rest of the manuscript.

\subsubsection{Clip tuples}
\label{sec:cliptuples}

Besides the case where each clip feature is matched independently, Perret \etal \cite{trx} additionally propose matching feature \emph{tuples}. Inspired by this, we extend Chamfer to enable matching of \emph{clip} feature tuples formed by any clip subset of fixed length $l$.
A clip feature tuple $\vt^{l}$ contains $l$ clip features, non-necessarily consecutive, but with the same relative order as in $Q = \{\vq_i\}$. For example, $\vt^{2} = \{ (q_i, q_{j \neq i)}\}$. 
Each clip feature tuple is concatenated and fed to the learnable projection head $\phi: \real^{ld} \rightarrow \real^D$. 
The resulting temporal similarity matrix is $M^{++}\in \real^{n'\times kn'}$, with $n'={n \choose l}$. 
The clip tuples are sub-sequence representations on top of single-clip representations.  By definition, clip tuples add additional temporal information to the representation. But the matching function remains non-temporal.

We also define non-temporally ordered clip tuples $\vt_{all}^{l}$ as the permutations of $l$ clip features. For example, $\vt_{all}^{2} = \{ (q_i, q_{j \neq i)}\}$. The resulting temporal similarity matrix is $M^{++}\in \real^{n'\times kn'}$ with $n'=n!$. 
The comparison between ordered and all tuples is presented in Table~\ref{tab:chamfer_ablation} and the appendix.
In the following, the extension of Chamfer matching that jointly matches multiple shots and uses clip tuples is referred to as \textbf{Chamfer++}. Unless otherwise stated, we use ordered clip tuples.

\section{Experiments}
\label{sec:experiments}

We report results on the three most commonly used benchmarks for few-shot action recognition, \ie \kinetics~\cite{zhu2018cmn}, Something-Something V2 (\ssvtwo) \cite{ssv2}, and \ucf \cite{ucf101}. 
\kinetics and \ucf contain videos collected from YouTube. Each video is trimmed to include only one coarse-grained human action such as ``playing trumpet'' or ``reading book''. \ssvtwo  contains egocentric videos where humans were instructed to perform predefined actions such as ``pushing something from left to right'' or ``dropping something into something''. 

We use the train/val/test splits from~\cite{zhu2018cmn} for all three datasets, containing 64/12/24 classes, respectively. 
We learn the parameters of the R(2+1)D backbone using the train split, similar to~\cite{tsl}. 
For matching-based approaches, we learn the projection $\phi$ together with any learnable parameters in the matching function $f$ using episodic training also on the train split. We use the val split for hyper-parameter tuning and early stopping.

\looseness=-1
We evaluate on the common 1-shot and 5-shot setups.
Unless otherwise stated, we use 5-way classification tasks.
Training episodes are randomly sampled from the train set.
We use the same fixed, predefined set of 10k test episodes sampled from the test set for all methods 
(prior work samples them randomly). 
We always evaluate \textit{three} trained models and report mean and standard deviation.

\subsection{Implementation details}
\label{sec:implementationdetails}
We use the publicly available code for TSL\footnote{\url{https://github.com/xianyongqin/few-shot-video-classification}} for 
training the backbone as well as for reproducing the TSL method. We adapt the public TRX\footnote{\url{https://github.com/tobyperrett/trx}} codebase for learning parameters of the matching function and testing,
as well as for reproducing the TRX and OTAM methods. 

\begin{table*}[t]
\vspace{-3ex}
\caption{\textbf{Few-shot action recognition results}. All methods are trained and evaluated by us and use the same R(2+1)D~\cite{r2plus1d} backbone. TSL~\cite{tsl} learns a classifier at each episode during testing. All the rest are pairwise matching-based methods and are split into two categories: non-temporal, \ie invariant to the temporal order of features, and temporal. $^\dagger$ denotes Chamfer++ matching using \textit{all} tuples.
Best (second-best) results are presented in \textbf{bold} (\underline{underlined}). 
\label{tab:main}
}
\centering
\resizebox{\linewidth}{!}
{
\setlength{\tabcolsep}{5pt}
\begin{tabular}{ l  ll ll ll} 
\toprule
     \multirow{2}{*}{Method}    & \multicolumn{2}{c}{\ssvtwo} & \multicolumn{2}{c}{\kinetics} & \multicolumn{2}{c}{\ucf}  \\
      & \oneshot & \fiveshot & \oneshot & \fiveshot & \oneshot & \fiveshot \\
     \midrule
      \multicolumn{5}{l}{~~\textit{Parametric classifier}} \vspace{2pt} \\
     TSL~\cite{tsl} & 60.6 \stddev{0.1} & 79.9 \stddev{0.0}  & 93.6 \stddev{0.0} & 98.0 \stddev{0.00}  & 97.1 \stddev{0.0} & \underline{99.4} \stddev{0.0}  \\ 
     \midrule
     \multicolumn{5}{l}{~~\textit{Non-temporal matching}} \vspace{2pt} \\
     Mean   & 65.8 \stddev{0.0} & 79.1 \stddev{0.1} & 95.5 \stddev{0.0} & 98.1 \stddev{0.1}  & 97.6 \stddev{0.2} & 98.9 \stddev{0.1}  \\
     Max  & 65.0  \stddev{0.2} & 79.0 \stddev{0.0} & 95.3 \stddev{0.1} & \underline{98.3} \stddev{0.0} & \textbf{97.9} \stddev{0.1} & 98.9 \stddev{0.0}  \\
     Chamfer++$^\dagger$ &  67.0 \stddev{0.3} & 80.8 \stddev{0.1} & \textbf{96.2} \stddev{0.1} & \textbf{98.4} \stddev{0.1} & \underline{97.8} \stddev{0.1} & 99.2 \stddev{0.1}   \\
     Chamfer++ &  \textbf{67.8} \stddev{0.2} & \underline{81.6 \stddev{0.1}} & \underline{96.1} \stddev{0.1} & \underline{98.3} \stddev{0.0} & 97.7 \stddev{0.0} & 99.3 \stddev{0.0}   \\
     
     \midrule
     \multicolumn{5}{l}{~~\textit{Temporal matching}} \vspace{2pt} \\
     Diagonal  & 66.7 \stddev{0.1} & 80.1 \stddev{0.0} & 95.3 \stddev{0.1} & 98.1 \stddev{0.1} & 97.6 \stddev{0.2} & 99.0 \stddev{0.0}  \\
     Linear  & 66.6 \stddev{0.1} & 80.1 \stddev{0.2}  & 95.5 \stddev{0.1} & 98.1 \stddev{0.0} & 97.6 \stddev{0.1} & 98.9 \stddev{0.0}   \\
     OTAM~\cite{otam} & 67.1 \stddev{0.0} & 80.2 \stddev{0.2} & 95.9 \stddev{0.0} & \textbf{98.4} \stddev{0.1} & \underline{97.8} \stddev{0.1} & 99.0 \stddev{0.0}  \\
     TRX-\{2,3\}~\cite{trx} & 65.5 \stddev{0.1} & \textbf{81.8} \stddev{0.2} & 93.4 \stddev{0.2} & 97.5 \stddev{0.0}  & 96.6 \stddev{0.0} & \textbf{99.5} \stddev{0.0}  \\
    ViSiL~\cite{visil} & \underline{67.7 \stddev{0.0}} & 81.3 \stddev{0.0} & 95.9 \stddev{0.0} & 98.2 \stddev{0.0}  & \underline{97.8} \stddev{0.2} & 99.0 \stddev{0.1}\\
     
\bottomrule
\end{tabular}
    }

\vspace{-2ex}

\end{table*}

\noindent\textbf{Learning the backbone parameters.}
We start from the publicly available 34-layer R(2+1)D backbone provided by the TSL codebase. This model is pre-trained on the large Sports-1M dataset~\cite{karpathy2014large}. We follow~\cite{tsl} and use a SGD optimizer with a constant learning rate of 0.001 for the backbone and 0.1 for the 64-class linear layer. We perform early-stopping using the 64-class validation dataset. 
Then, we use the backbone as a feature extractor to extract features from $n=8$ uniformly sampled clips\footnote{Note that although TSL uses randomly-sampled clips, we found that its performance is usually better when switching to uniformly-sampled clips.}. The input video clips are composed of 16 consecutive RGB frames with a spatial resolution of 112x112, and the dimensionality of the resulting feature vector is $d=512$.

\noindent\textbf{Training matching-based methods.}
We train the matching-based methods on the training episodes with an SGD optimizer and a constant learning rate of 0.001 for every method except for TRX where we use a learning rate of 0.01. 
Similar to prior work~\cite{otam,trx}, we select the best model using early stopping by measuring performance on the
validation set. We learn the projection $\phi$ jointly with any matching parameters. We set the projection dimension to $D=1152$ if not stated otherwise. This is equivalent to the dimensionality that TRX\cite{trx} uses for its attention layer. We train all the matching methods with the cross-entropy loss that uses softmax with a learnable temperature $\tau$.

\noindent\textbf{Learning classifiers for TSL.}
Instead of pairwise matching, TSL~\cite{tsl} learns classifiers at every test episode, using all available support examples. To reproduce TSL we
follow~\cite{tsl} and use the Adam optimizer with a constant learning rate of 0.01 for 10 epochs. 
The original TSL approach uses $n=10$ clips at test-time, but we set this number to $n=8$ to keep it the same with all matching methods for a fair comparison. Preliminary experiments show that this choice doesn't affect the performance of TSL at all.

\noindent\textbf{Data augmentation.} During 
training, videos are augmented with random cropping. 
We uniformly sample 8 clips from each video with temporal jittering, \ie  randomly perturb the starting point of each clip.
Additionally, for the \kinetics dataset, we also use random horizontal flipping as data augmentation. 
Since it is important for \ssvtwo to distinguish between
left-to-right and right-to-left, we do not use horizontal flipping for that dataset.
We only apply a center crop for videos during validation and testing.

\subsection{Results}
\label{sec:results}
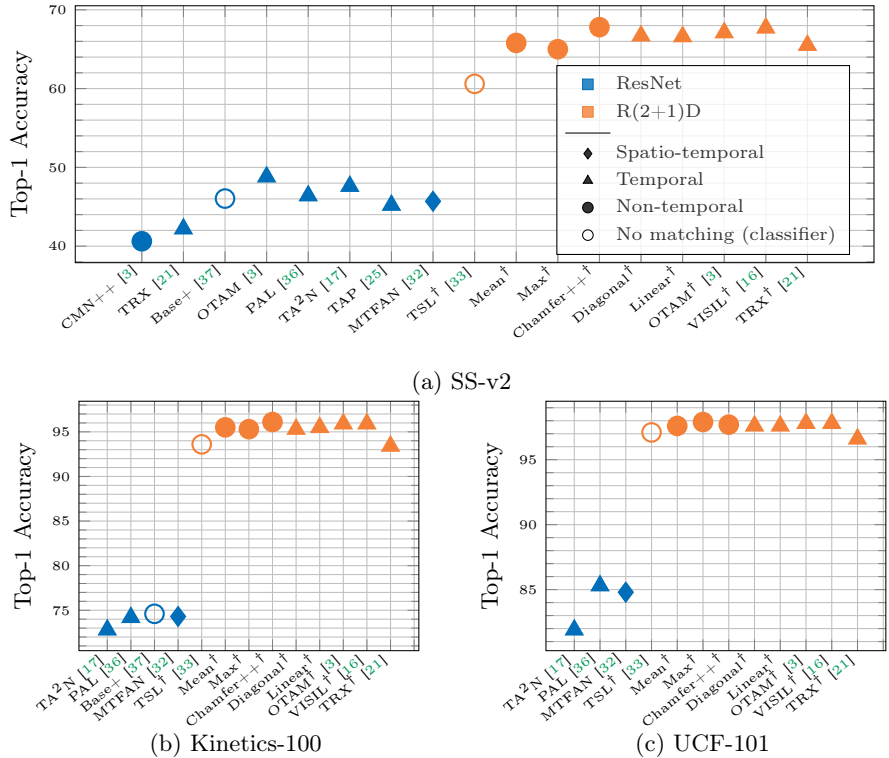
\begin{figure}[t]
\vspace{-3ex}
\centering

\begin{subfigure}{\textwidth}
\begin{tikzpicture}
\begin{axis}[
  width=\linewidth,
  height=5cm,
  xtick = {0,1,2,3,4,5,6,7,8,9,10,11,12, 13, 14, 15, 16},
  xticklabels = {CMN++~\cite{otam},TRX~\cite{trx},Base+~\cite{Zhu2021CloserLook},OTAM~\cite{otam},  PAL \cite{Zhu2021pal}, TA$^2$N~\cite{li2021ta2n}, TAP~\cite{su2022temporal}, MTFAN~\cite{Wu_2022_CVPR}, TSL$^\dagger$~\cite{tsl},  Mean$^\dagger$, Max$^\dagger$, Chamfer++$^\dagger$, Diagonal$^\dagger$, Linear$^\dagger$, OTAM$^\dagger$~\cite{otam}, VISIL$^\dagger$~\cite{visil}, TRX$^\dagger$~\cite{trx}},
  x tick label style={rotate=40,anchor=east},
  ylabel={\small Top-1 Accuracy},
  legend pos=south east,
  label style={font=\scriptsize, row sep=0.5pt},
  tick label style={font=\tiny},
  ylabel near ticks, xlabel near ticks, 
  legend style={font=\scriptsize}, 
  minor y tick num=4,
  ]

    \addlegendimage{color=\resnetc,mark=square*,only marks} \addlegendentry{ResNet};
    \addlegendimage{color=\rtwoplusonec,mark=square*,only marks} \addlegendentry{R(2+1)D};
    \addlegendimage{color=black,mark=} \addlegendentry{};
    \addlegendimage{color=black,mark=\markerspatiotemporal,only marks}
    \addlegendentry{Spatio-temporal};
    \addlegendimage{color=black,mark=\markertemporal,only marks} \addlegendentry{Temporal};
    \addlegendimage{color=black,mark=\markernontemporal,only marks} \addlegendentry{Non-temporal};
    \addlegendimage{color=black,mark=o,only marks} \addlegendentry{No matching (classifier)};
    
    \addplot[cmn] coordinates {(0, 40.62)};
    \addplot[trxresnet] coordinates {(1, 42.2)};
    \addplot[bmvc] coordinates {(2, 46.04)};
    \addplot[otamresnet] coordinates {(3, 48.8)};
    \addplot[pal] coordinates {(4, 46.4)};
    \addplot[titan] coordinates {(5, 47.6)};
    \addplot[tap] coordinates {(6, 45.2)};
    \addplot[mtfan] coordinates {(7, 45.7)};
    
    \addplot[tsl] coordinates {(8, 60.6)};
    \addplot[mean] coordinates {(9, 65.8)};
    \addplot[max] coordinates {(10, 65.0)};
    \addplot[chamfer] coordinates {(11, 67.8)};
    \addplot[diag] coordinates {(12, 66.7)};
    \addplot[linear] coordinates {(13, 66.6)};
    \addplot[otam] coordinates {(14, 67.1)};
    \addplot[visil] coordinates {(15, 67.7)};
    \addplot[trx] coordinates {(16, 65.5)};
            
\end{axis}
\end{tikzpicture} 
\caption{\ssvtwo}
\label{fig:teaser_ssv2}
\end{subfigure}  

\begin{tabular}{c}

\begin{subfigure}{0.5\textwidth}
\begin{tikzpicture}
\begin{axis}[
  height=4.9cm,
  xtick = {0,1,2,3,4,5,6,7,8,9,10,11,12, 13, 14, 15, 16},
  xticklabels = {TA$^2$N~\cite{li2021ta2n}, PAL \cite{Zhu2021pal}, Base+~\cite{Zhu2021CloserLook},   MTFAN~\cite{Wu_2022_CVPR}, TSL$^\dagger$~\cite{tsl},  Mean$^\dagger$, Max$^\dagger$, Chamfer++$^\dagger$, Diagonal$^\dagger$, Linear$^\dagger$, OTAM$^\dagger$~\cite{otam}, VISIL$^\dagger$~\cite{visil}, TRX$^\dagger$~\cite{trx}},
  x tick label style={rotate=40,anchor=east},
  ylabel={\small Top-1 Accuracy},
  legend pos=south east,
  label style={font=\tiny, row sep=0.5pt},
  tick label style={font=\tiny},
  ylabel near ticks, xlabel near ticks, 
  legend style={font=\scriptsize}, 
  minor y tick num=4,
  ]

    \addplot[titan] coordinates {(0, 72.8)};
    \addplot[pal] coordinates {(1, 74.2)};
    \addplot[bmvc] coordinates {(2, 74.6)};
    \addplot[mtfan] coordinates {(3, 74.3)};
    
    \addplot[tsl] coordinates {(4, 93.6)};
    \addplot[mean] coordinates {(5, 95.5)};
    \addplot[max] coordinates {(6, 95.3)};
    \addplot[chamfer] coordinates {(7, 96.1)};
    \addplot[diag] coordinates {(8, 95.3)};
    \addplot[linear] coordinates {(9, 95.5)};
    \addplot[otam] coordinates {(10, 95.9)};
    \addplot[visil] coordinates {(11, 95.9)};
    \addplot[trx] coordinates {(12, 93.4)};

\end{axis}
\end{tikzpicture} 
 \vspace{-10pt}
\caption{\kinetics}
\label{fig:teaser_kinetics}
\end{subfigure}  
\begin{subfigure}{0.5\textwidth}
\begin{tikzpicture}
\begin{axis}[
  width=\linewidth,
  height=4.9cm,
  xtick = {0,1,2,3,4,5,6,7,8,9,10,11,12},
  xticklabels = {TA$^2$N~\cite{li2021ta2n}, PAL \cite{Zhu2021pal}, MTFAN~\cite{Wu_2022_CVPR}, TSL$^\dagger$~\cite{tsl}, Mean$^\dagger$, Max$^\dagger$, Chamfer++$^\dagger$, Diagonal$^\dagger$, Linear$^\dagger$, OTAM$^\dagger$~\cite{otam}, VISIL$^\dagger$~\cite{visil},TRX$^\dagger$~\cite{trx}},
  x tick label style={rotate=40,anchor=east},
  ylabel={\small Top-1 Accuracy},
  legend pos=south east,
  label style={font=\scriptsize, row sep=0.5pt},
  tick label style={font=\tiny},
  ylabel near ticks, xlabel near ticks, 
  legend style={font=\scriptsize}, 
  minor y tick num=4,
  ]

    \addplot[titan] coordinates {(0, 81.9)};
    \addplot[pal] coordinates {(1, 85.3)};
    \addplot[mtfan] coordinates {(2, 84.8)};
    
    \addplot[tsl] coordinates {(3, 97.1)};
    \addplot[mean] coordinates {(4, 97.6)};
    \addplot[max] coordinates {(5, 97.9)};
    \addplot[chamfer] coordinates {(6, 97.7)};
    \addplot[diag] coordinates {(7, 97.6)};
    \addplot[linear] coordinates {(8, 97.6)};
    \addplot[otam] coordinates {(9, 97.8)};
    \addplot[visil] coordinates {(10, 97.8)};
    \addplot[trx] coordinates {(11, 96.6)};
    
\end{axis}
\end{tikzpicture} 
 \vspace{-10pt}
\caption{\ucf}
\label{fig:teaser_ucf101}
\end{subfigure}  

\end{tabular}
\caption{\textbf{One-shot performance} for different backbones, types of matching, or use of parametric classifiers. The different colors account for the different backbones. The different shapes account for the type of matching.
$^\dagger$ denotes methods reproduced in this study.}
\label{fig:teasers}
\end{figure}

In this section, we report and analyse our results. We first discuss the gains from using temporal representations and the comparison between matching-based and classifier approaches. We then discuss the use of temporal information during matching and present results when varying the number of classes per classification task (test episode). Finally, we compare Chamfer++ to other recently published methods and show that our approach achieves state-of-the-art performance.

\myparagraph{Frame or clip-based features?} 
We start by evaluating a number of recent matching-based methods over temporal features. This is an important comparison that is missing from the current few-shot action recognition literature.
In Figure~\ref{fig:teasers}, we report one-shot performance for a number of matching methods under a common evaluation setup and using both frame-based (ResNet, blue points) and clip-based (R(2+1)D, orange points) features.

We clearly see that using a spatio-temporal backbone significantly boosts accuracy by more than 10\% on all datasets. 
Interestingly, this is also true for the \kinetics and the \ucf datasets that are known to be more biased towards spatial context~\cite{huang2018makes}. Even for this case where context is important, we see that temporal dynamics
remain a valuable cue for few-shot action recognition. 
It is worth noting that the performance on \ucf and \kinetics appears to be saturated when using spatio-temporal representations.

\myparagraph{Pairwise matching or classifiers?}
Matching-based methods and 
classifiers are compared in Table~\ref{tab:main} and Figure~\ref{fig:teasers}.
All results in the table are computed under a common framework, \ie all methods share representations from an R(2+1)D backbone and are tested on the same set of episodes. We see that for
all setups and datasets, several matching approaches outperform TSL. In the 1-shot regime, \textit{most} matching-based methods outperform TSL.

\begin{table*}[t]
\vspace{-15pt}  
\caption{\textbf{Comparison with the state-of-the-art.} Unless otherwise stated, we report results as presented in the corresponding papers. $^\dagger$ denotes results reproduced in \cite{Zhu2021CloserLook} and $^\ddagger$ denotes results generated by us. Best (second-best) results are presented in \textbf{bold} (\underline{underlined}).
\label{tab:sota-papers}
}
\centering        

\setlength\extrarowheight{-1pt}
    \resizebox{\linewidth}{!}
    {
    \begin{tabular}{ lcccccccc } 
    \toprule
        \multirow{2}{*}{Method}  &  Clip-based & \multirow{2}{*}{Venue} & \multicolumn{2}{c}{\ssvtwo}  & \multicolumn{2}{c}{\kinetics} & \multicolumn{2}{c}{\ucf}\\
         & backbone&  &  \oneshot & \fiveshot & \oneshot & \fiveshot & \oneshot & \fiveshot \\
         \midrule
         \multicolumn{5}{l}{~~\textit{Classifier-based}} \vspace{2pt} \\
        Baseline \cite{Zhu2021CloserLook} &  & BMVC21 & 40.8 & 59.2 & 69.5 & 84.4 & - & -   \\
        Baseline + \cite{Zhu2021CloserLook} &   &  BMVC21 & 46.0 & 61.1 & 74.6 & 86.6 & - & -   \\
            
        TSL~\cite{tsl} & \checkmark & PAMI21 &59.1& \underline{80.1} & 92.5 & 97.8 & 94.8 & -  \\
        TSL $^\ddagger$ & \checkmark & PAMI21 & \underline{60.6} & 79.9 & \underline{93.6} & \underline{98.0} & 97.1 & \textbf{99.4}  \\
        \midrule
        \multicolumn{5}{l}{~~\textit{Matching-based}} \vspace{2pt} \\
        CMN++ $^\dagger$ \cite{zhu2018cmn,Zhu2021CloserLook} &  & ECCV18 & 40.6  & 51.9 & 65.9 & 82.7 & - & -   \\
        ARN~\cite{Zhang2020FewShotAR} & \checkmark & ECCV20 & - & - & 63.7 & 82.4 & 66.3 & 83.1   \\
        OTAM \cite{otam} &  & CVPR20 & 48.8 & 52.3 & 73.0 & 85.8 & - & -    \\
        PAL \cite{Zhu2021pal} &  &  BMVC21 & 46.4 & 62.6 & 74.2 & 87.1 & 85.3 & 95.2   \\ 
        TRX-\{1\}~\cite{trx} &    &  CVPR21 & 38.8 & 60.6 & 63.6 & 85.2 & - & -   \\
        TRX-\{2,3\}~\cite{trx} &  &  CVPR21 & 42.0 & 64.6 & 63.6 & 85.9 & - & 96.1   \\
        STRM-\{2\}~\cite{thatipelli2021spatio} &   & CVPR22 & - & 70.2 &  - & 91.2 & - & 98.1  \\
        TA$^2$N~\cite{li2021ta2n} & & AAAI22 & 47.6 & 61.0 & 72.8 & 85.8 & 81.9 & 95.1   \\
        MTFAN ~\cite{Wu_2022_CVPR} &  & CVPR22 & 45.7  & 60.4 &  74.6 & 87.4 & 84.8 & 95.1  \\
        HyRSM~\cite{Wang_2022_CVPR} &  & CVPR22 & 54.3 & 69.0 & 73.7 & 86.1 & - & -  \\
        TAP  ~\cite{su2022temporal}    & & CVPR22 & 45.2 & 63.0 & - & - & 83.9 & 95.4  \\
        CPM ~\cite{Yang_2022_ECCV}  &  & ECCV22 & 59.6 & - & 81.0 &- & 79.0 & -  \\

         Chamfer++ 
         $^\ddagger$ & \checkmark & & \textbf{67.8} & \textbf{81.6} & \textbf{96.1}  & \textbf{98.3}  & \textbf{97.7} & 99.3 \\
    \bottomrule
    \end{tabular}
        }

\vspace{-3.0ex}
\end{table*}

\myparagraph{How useful is temporal matching?}
No significant difference in performance is observed between temporal and non-temporal matching approaches, as highlighted in Table~\ref{tab:main} and Figure~\ref{fig:teasers}.
On the \kinetics and \ucf datasets, where action classes are generally coarser and highly dependent on context~\cite{huang2018makes}, most methods we tested perform similarly well. Simply using $f(M)= \max_{ij} m_{ij}$ is enough to achieve top performance for \ucf, while the proposed Chamfer++ outperforms all other methods on \kinetics.
When it comes to the finer-grained \ssvtwo dataset, we see that, although most non-temporal matching methods lag behind the temporal ones for the 1-shot case, the proposed Chamfer++ method achieves the highest performance without temporal matching. For 5-shot action recognition, TRX, ViSiL, and Chamfer++ perform similarly well, with the last being parameter-free, more intuitive, and faster.

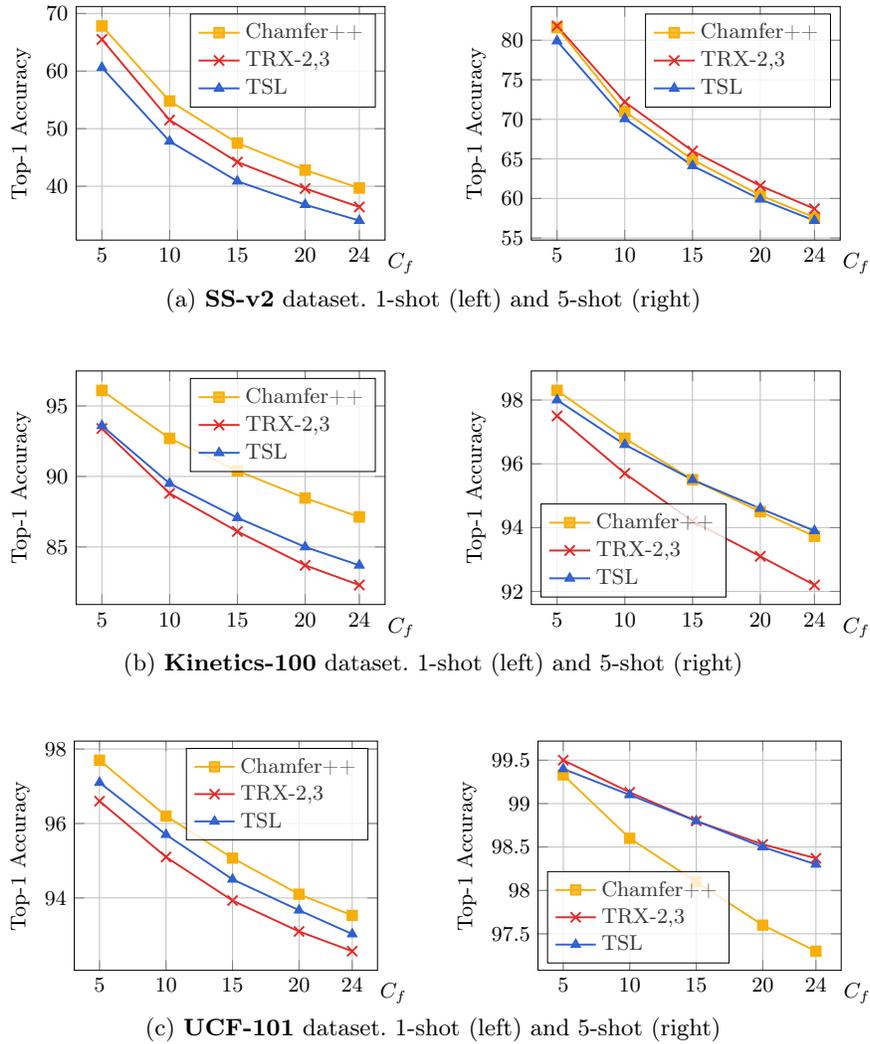
\begin{figure*}[t!]
\centering
 \begin{subfigure}{\linewidth}
    \resizebox{\linewidth}{!}{
      \definecolor{C1}{RGB}{226, 43, 41}
\definecolor{C2}{RGB}{47, 96, 206}
\definecolor{C3}{RGB}{246, 175, 11} 

\begin{tikzpicture}
  \begin{axis}[
    width=0.5\linewidth,
    height=5cm, 
    xtick = {5,10,15,20,24},
	legend cell align={left},
	legend pos=north east,
    grid=both,
    ylabel={\small Top-1 Accuracy},
    xlabel={\small $C_f$},
    xlabel style={yshift=10pt, xshift=80pt,anchor=north east},
  ]
    
\addplot [thick, color=C3, mark=square*,  mark size=2] coordinates {(5, 67.83) (10, 54.8) (15, 47.5) (20, 42.8) (24, 39.7)};    
\addlegendentry{Chamfer++} 

\addplot [thick, color=C1, mark=x,  mark size=3] coordinates {(5, 65.5) (10, 51.5) (15, 44.2) (20, 39.6) (24, 36.4)};      
\addlegendentry{TRX-{2,3}} 
	
\addplot [thick, color=C2, mark=triangle*,  mark size=2] coordinates {(5, 60.6) (10, 47.8) (15, 40.87) (20, 36.8) (24, 34.03)};      
\addlegendentry{TSL} 
    
  \end{axis}
\end{tikzpicture}
      \definecolor{C1}{RGB}{226, 43, 41}
\definecolor{C2}{RGB}{47, 96, 206}
\definecolor{C3}{RGB}{246, 175, 11}

\begin{tikzpicture}
  \begin{axis}[
    width=0.5\linewidth,
    height=5cm, 
    xtick = {5,10,15,20,24},
	legend cell align={left},
	legend pos=north east,
    grid=both,
    ylabel={\small Top-1 Accuracy},
    xlabel={\small $C_f$},
    xlabel style={yshift=10pt, xshift=80pt,anchor=north east},
  ]

\addplot [thick, color=C3, mark=square*,  mark size=2] coordinates {(5, 81.60) (10, 71) (15, 64.9) (20, 60.4) (24, 57.6)};   
\addlegendentry{Chamfer++} 

\addplot [thick, color=C1, mark=x,  mark size=3] coordinates {(5, 81.8) (10, 72.2) (15, 66) (20, 61.6) (24, 58.7)};     
\addlegendentry{TRX-{2,3}} 
	
\addplot [thick, color=C2, mark=triangle*,  mark size=2] coordinates {(5, 79.9) (10, 70.07) (15, 64.13) (20, 59.9) (24, 57.2)};  
\addlegendentry{TSL} 
    
  \end{axis}
\end{tikzpicture}
    }
    \vspace{-15pt}
    \caption{\textbf{\ssvtwo} dataset. 1-shot (left) and  5-shot (right)}
    \label{fig:ways_ssv2}
\end{subfigure}     

\vspace{20pt}
\begin{subfigure}{\linewidth}
    \resizebox{\linewidth}{!}{
      \definecolor{C1}{RGB}{226, 43, 41}
\definecolor{C2}{RGB}{47, 96, 206}
\definecolor{C3}{RGB}{246, 175, 11} 

\begin{tikzpicture}
  \begin{axis}[
    width=0.5\linewidth,
    height=5cm, 
    xtick = {5,10,15,20,24},
	legend cell align={left},
	legend pos=north east,
    grid=both,
    ylabel={\small Top-1 Accuracy},
    xlabel={\small $C_f$},
    xlabel style={yshift=10pt, xshift=80pt,anchor=north east},
  ]
    
    \addplot [thick, color=C3, mark=square*,  mark size=2] coordinates {(5, 96.1) (10, 92.7) (15, 90.4) (20, 88.47) (24, 87.13)};     
    \addlegendentry{Chamfer++}  

    \addplot [thick, color=C1, mark=x,  mark size=3] coordinates {(5, 93.4) (10, 88.8) (15, 86.1) (20, 83.7) (24, 82.3)};      
    \addlegendentry{TRX-{2,3}} 
	
    \addplot [thick, color=C2, mark=triangle*,  mark size=2] coordinates {(5, 93.6) (10, 89.5) (15, 87.07) (20, 85) (24, 83.7)};      
    \addlegendentry{TSL}

  \end{axis}
\end{tikzpicture}
      \definecolor{C1}{RGB}{226, 43, 41}
\definecolor{C2}{RGB}{47, 96, 206}
\definecolor{C3}{RGB}{246, 175, 11}

\begin{tikzpicture}
  \begin{axis}[
    width=0.5\linewidth,
    height=5cm, 
    xtick = {5,10,15,20,24},
	legend cell align={left},
	legend pos=south west,
    grid=both,
    ylabel={\small Top-1 Accuracy},
    xlabel={\small $C_f$},
    xlabel style={yshift=10pt, xshift=80pt,anchor=north east},
  ]

    \addplot [thick, color=C3, mark=square*,  mark size=2] coordinates {(5, 98.3) (10, 96.8) (15, 95.5) (20, 94.5) (24, 93.73)};   
    \addlegendentry{Chamfer++}  
	
    \addplot [thick, color=C1, mark=x,  mark size=3] coordinates {(5, 97.5) (10, 95.7) (15, 94.2) (20, 93.1) (24, 92.2)};      
    \addlegendentry{TRX-{2,3}} 
	
    \addplot [thick, color=C2, mark=triangle*,  mark size=2] coordinates {(5, 98) (10, 96.6) (15, 95.5) (20, 94.6) (24, 93.9)};      
    \addlegendentry{TSL}

  \end{axis}
\end{tikzpicture}
    }
    \vspace{-15pt}
    \caption{\textbf{\kinetics} dataset. 1-shot (left) and  5-shot (right)}
    \label{fig:ways_kinetics}
\end{subfigure}     

\vspace{20pt}
\begin{subfigure}{\linewidth}
    \resizebox{\linewidth}{!}{
      \definecolor{C1}{RGB}{226, 43, 41}
\definecolor{C2}{RGB}{47, 96, 206}
\definecolor{C3}{RGB}{246, 175, 11}

\begin{tikzpicture}
  \begin{axis}[
    width=0.5\linewidth,
    height=5cm, 
    xtick = {5,10,15,20,24},
	legend cell align={left},
	legend pos=north east,
    grid=both,
    ylabel={\small Top-1 Accuracy},
    xlabel={\small $C_f$},
    xlabel style={yshift=10pt, xshift=80pt,anchor=north east},
  ]

\addplot [thick, color=C3, mark=square*,  mark size=2] coordinates {(5, 97.7)(10, 96.2) (15, 95.07) (20, 94.1) (24, 93.53)}; 
\addlegendentry{Chamfer++} 

\addplot [thick, color=C1, mark=x,  mark size=3] coordinates {(5, 96.6) (10, 95.1) (15, 93.93) (20, 93.1) (24, 92.57)};       
\addlegendentry{TRX-{2,3}} 
	
\addplot [thick, color=C2, mark=triangle*,  mark size=2] coordinates {(5, 97.1) (10, 95.7) (15, 94.5) (20, 93.67) (24, 93.03)};   
\addlegendentry{TSL} 

  \end{axis}
\end{tikzpicture}
      \definecolor{C1}{RGB}{226, 43, 41}
\definecolor{C2}{RGB}{47, 96, 206}
\definecolor{C3}{RGB}{246, 175, 11}

\begin{tikzpicture}
  \begin{axis}[
    width=0.5\linewidth,
    height=5cm, 
    xtick = {5,10,15,20,24},
	legend cell align={left},
	legend pos=south west,
    grid=both,
    ylabel={\small Top-1 Accuracy},
    xlabel={\small $C_f$},
    xlabel style={yshift=10pt, xshift=80pt,anchor=north east},
  ]
\addplot [thick, color=C3, mark=square*,  mark size=2] coordinates {(5, 99.33)(10, 98.6)(15, 98.1)(20, 97.6) (24, 97.3)};
\addlegendentry{Chamfer++} 

\addplot [thick, color=C1, mark=x,  mark size=3] coordinates {(5, 99.5) (10, 99.13) (15, 98.8) (20, 98.53) (24, 98.37)};   
\addlegendentry{TRX-{2,3}} 

\addplot [thick, color=C2, mark=triangle*,  mark size=2] coordinates {(5, 99.4) (10, 99.1) (15, 98.8) (20, 98.5) (24, 98.3)};     
\addlegendentry{TSL} 

  \end{axis}
\end{tikzpicture}
    }
    \vspace{-15pt}
    \caption{\textbf{\ucf} dataset. 1-shot (left) and  5-shot (right)}
    \label{fig:ways_ucf}
\end{subfigure}     

\caption{\textbf{Impact of the number of classes  per episode ($C_{f}$)} on three datasets.
\label{fig:number_ways}
}
\vspace{-7ex}
\end{figure*}
\myparagraph{Varying the number of classes in an episode.}
Figure~\ref{fig:number_ways} shows the performance when extending the case of $C_{f}=5$ classes to the maximum number of classes, $C_{f}=24$, in the 1-shot and 5-shot regime. 
We see that for 1-shot the proposed non-temporal Chamfer++ method highly outperforms TRX and the classifier-based TSL method in all datasets.

\begin{wraptable}[11]{R}{0.5\linewidth}
\vspace{-20pt}
\caption{Chamfer++ variants.} 
\label{tab:chamfer_ablation}
\centering
    \resizebox*{!}{95pt}
    {
        \setlength\extrarowheight{-3pt}
    \begin{tabular}{@{\xssp} l@{\ssp} l@{\ssp}l@{\ssp} l@{\ssp}l@{\ssp}l@{\nsp}} 
    \toprule
         Method & $l$ & {\scriptsize Joint} & {\scriptsize Tupl.} & \oneshot & \fiveshot \\
         \midrule
         Chamfer-Q  &  1 &   & &  65.7 \hspace{-3pt}\stddev{0.1} & 79.7 \hspace{-3pt}\stddev{0.1} \\
         Chamfer-S  &  1 &   & &  65.3 \hspace{-3pt}\stddev{0.1} & 79.1 \hspace{-3pt}\stddev{0.2}  \\
         Chamfer  &  1 &     & & 66.9 \hspace{-3pt}\stddev{0.1} & 80.0 \hspace{-3pt}\stddev{0.2} \\
         Chamfer+  &  1 & \checkmark   &  & 66.9 \hspace{-3pt}\stddev{0.1}& 80.7 \hspace{-3pt}\stddev{0.2} \\

         Chamfer++  &  2 &  \checkmark  & all & 67.1 \hspace{-3pt}\stddev{0.1} &  80.8 \hspace{-3pt}\stddev{0.2} \\
         Chamfer++  &  2 &  \checkmark  &  ord. & 67.7 \hspace{-3pt}\stddev{0.1} & 81.4 \hspace{-3pt}\stddev{0.2} \\
         Chamfer++  &  3 &  \checkmark  & all & 67.0 \hspace{-3pt}\stddev{0.3} & 80.8 \hspace{-3pt}\stddev{0.1} \\
         Chamfer++  &  3 &  \checkmark  & ord. & 67.8 \hspace{-3pt}\stddev{0.2} & 81.6 \hspace{-3pt}\stddev{0.1} \\
    \bottomrule
    \end{tabular}
        
    }
\vspace{-2ex}

\vspace{-3.5ex}
\end{wraptable}

\myparagraph{Comparison to the state-of-the-art.}
In Table~\ref{tab:sota-papers}, we compare the performance of the proposed Chamfer++ with the corresponding numbers reported in many recent few-shot action recognition papers. Although there is no direct comparison between all these methods, \ie, no common setup or backbones, we present all results jointly to show the overall progress in the task.

\subsection{Chamfer matching ablation and interpretability}
\label{sec:ablation}

In Table~\ref{tab:chamfer_ablation}, we present an ablative study with the 1-shot performance of all the different variants of Chamfer++ we discuss in Section~\ref{sec:newchamfer} on the \ssvtwo dataset. Combining support-based and query-based Chamfer matching helps learn a better projection $\phi$, while both joint matching and the use of clip tuples improve performance compared to the vanilla variant. Although using ordered clip tuples improves for 1-shot on \ssvtwo, overall, we see in Table~\ref{tab:main} that using all or ordered tuples results in more or less similar performance. 

\wrapfill

\begin{figure*}[t]
\vspace{-0.5ex}
\centering
\input{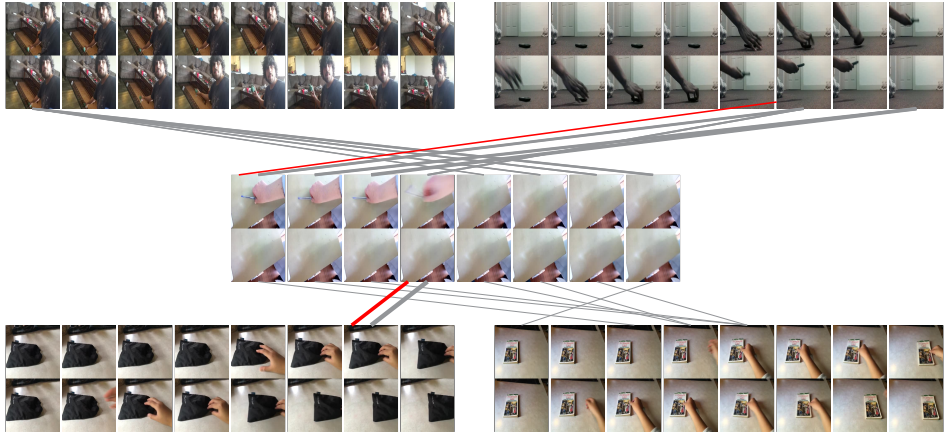}
\vspace{-2ex}
\caption{\textbf{Matching a query video (middle) from “Picking something up” with two support videos of “Picking something up” (top) and two support videos of “Pulling something from left to right” (bottom)}. 
Each video consists of 8 clips with 16 frames each. We only show the first and last frames on top of each other in the figure (see project page for animated versions).
In grey, we draw the correspondences between the query and the support videos selected by Chamfer (query-based). In red, we draw the correspondence selected by max (single strongest correspondence). 
Line thickness corresponds to the pairwise similarity.  
\label{fig:example}
\vspace{-3.5ex}}
\end{figure*}

\looseness=-1
\myparagraph{Which are the most informative clips?}
As we presented in Section~\ref{sec:preliminaries}, we sample and encode a set of clips to represent each video. Not all clips are, however, equally valuable for the matching process.
To provide some insight and illustrate the proposed method's interpretability, we study a qualitative example from \ssvtwo in Figure \ref{fig:example}.
When matching the query video to the two support videos belonging to the same class (top), we see many clip-to-clip correspondences with high magnitudes (shown as thicker lines). This is not the case when matching to the support videos belonging to a negative class (bottom) and we can only see a single strong correspondence. Upon better inspection, we saw that the motion of picking up the object in the fourth clip of the query video does exhibit a left-to-right motion locally that matches the negative class. Nevertheless, the other correspondences are weaker.
Leveraging more than one clip helps Chamfer matching to disambiguate while, at the same time, its inherent selectivity makes it more robust against noisy correspondences.
This example illustrates that not all the clip correspondences are equally valuable to compute a video-to-video similarity metric. The matching step is a dynamic way to select the most informative clip correspondences that highlight the similarity between two videos.

\section{Conclusion}
\label{sec:conclusion}

\vspace{-6pt}
A number of recent few-shot learning papers are abandoning the meta-learning protocol for representation learning~\cite{wang2019simpleshot,Zhu2021CloserLook,chen2020closer,tsl} and show that adaptation from the best possible features leads to better performance. 
In the quest for rapid adaptation at test time, we also adopt this setup and show that, given strong visual representations, simple matching-based methods are really effective and able to beat both classifier-based and more complex matching-based approaches on many common benchmarks for few-shot action recognition.
We further show that temporal information in the matching provides no particular benefit compared to the ability to learn or adapt from strong temporal features, and introduce an intuitive matching-based method that is not only parameter-free but also easy to visualize and interpret.

\noindent\textbf{Acknowledgements.}
This work was supported by Naver Labs Europe, by Junior Star GACR GM 21-28830M, and by student grant SGS23/173/OHK3/3T/13. The authors would like to sincerely thank Toby Perrett and Dima Damen for sharing their early code and supporting us, Diane Larlus for insightful conversations, feedback, and support, and Zakaria Laskar, Monish Keswani, and Assia Benbihi for their feedback.

{\small
\bibliographystyle{ieee_fullname}
\bibliography{egbib}

\begin{thebibliography}{10}\itemsep=-1pt

\bibitem{bishay2019tarn}
Mina Bishay, Georgios Zoumpourlis, and I. Patras.
\newblock Tarn: Temporal attentive relation network for few-shot and zero-shot
  action recognition.
\newblock In {\em BMVC}, 2019.

\bibitem{cao21comet}
Kaidi Cao, Maria Brbi\'c, and Jure Leskovec.
\newblock Concept learners for few-shot learning.
\newblock In {\em ICLR}, 2021.

\bibitem{otam}
Kaidi Cao, Jingwei Ji, Zhangjie Cao, Chien-Yi Chang, and Juan~Carlos Niebles.
\newblock Few-shot video classification via temporal alignment.
\newblock In {\em CVPR}, 2020.

\bibitem{chang2019d3tw}
Chien-Yi Chang, De-An Huang, Yanan Sui, Li Fei-Fei, and Juan~Carlos Niebles.
\newblock D3tw: Discriminative differentiable dynamic time warping for weakly
  supervised action alignment and segmentation.
\newblock In {\em CVPR}, 2019.

\bibitem{chen2020closer}
Wei-Yu Chen, Yen-Cheng Liu, Zsolt Kira, Yu-Chiang~Frank Wang, and Jia-Bin
  Huang.
\newblock A closer look at few-shot classification.
\newblock In {\em ICLR}, 2019.

\bibitem{Doersh2020CrossTransformers}
Carl Doersch, Ankush Gupta, and Andrew Zisserman.
\newblock Crosstransformers: spatially-aware few-shot transfer.
\newblock In {\em NeurIPS}, 2020.

\bibitem{dvornik2021drop}
Mikita Dvornik, Isma Hadji, Konstantinos~G Derpanis, Animesh Garg, and Allan
  Jepson.
\newblock Drop-dtw: Aligning common signal between sequences while dropping
  outliers.
\newblock In {\em NeurIPS}, 2021.

\bibitem{Dwivedi2019Protogan}
Sai~Kumar Dwivedi, Vikram Gupta, Rahul Mitra, Shuaib Ahmed, and Arjun Jain.
\newblock Protogan: Towards few shot learning for action recognition.
\newblock In {\em ICCVW}, 2019.

\bibitem{finn2017model}
Chelsea Finn, Pieter Abbeel, and Sergey Levine.
\newblock Model-agnostic meta-learning for fast adaptation of deep networks.
\newblock In {\em ICML}, 2017.

\bibitem{Finn2018}
Chelsea Finn, Kelvin Xu, and Sergey Levine.
\newblock Probabilistic model-agnostic meta-learning.
\newblock In {\em NeurIPS}, 2018.

\bibitem{GK18}
Spyros Gidaris and Nikos Komodakis.
\newblock Dynamic few-shot visual learning without forgetting.
\newblock In {\em CVPR}, 2018.

\bibitem{ssv2}
R. Goyal, S. Kahou, V. Michalski, J. Materzynska, S. Westphal, H. Kim, V.
  Haenel, I. Fruend, P. Yianilos, M. Mueller-Freitag, F. Hoppe, C. Thurau, I.
  Bax, and R. Memisevic.
\newblock The "something something" video database for learning and evaluating
  visual common sense.
\newblock In {\em ICCV}, 2017.

\bibitem{Grant2018}
Erin Grant, Chelsea Finn, Sergey Levine, Trevor Darrell, and Thomas Griffiths.
\newblock Recasting gradient-based meta-learning as hierarchical bayes.
\newblock In {\em ICLR}, 2018.

\bibitem{huang2018makes}
De-An Huang, Vignesh Ramanathan, Dhruv Mahajan, Lorenzo Torresani, Manohar
  Paluri, Li Fei-Fei, and Juan~Carlos Niebles.
\newblock What makes a video a video: Analyzing temporal information in video
  understanding models and datasets.
\newblock In {\em CVPR}, 2018.

\bibitem{karpathy2014large}
Andrej Karpathy, George Toderici, Sanketh Shetty, Thomas Leung, Rahul
  Sukthankar, and Li Fei-Fei.
\newblock Large-scale video classification with convolutional neural networks.
\newblock In {\em CVPR}, 2014.

\bibitem{visil}
Giorgos Kordopatis-Zilos, Symeon Papadopoulos, Ioannis Patras, and Ioannis
  Kompatsiaris.
\newblock Visil: Fine-grained spatio-temporal video similarity learning.
\newblock In {\em ICCV}, 2019.

\bibitem{li2021ta2n}
Shuyuan Li, Huabin Liu, Rui Qian, Yuxi Li, John See, Mengjuan Fei, Xiaoyuan Yu,
  and Weiyao Lin.
\newblock Ta2n: Two-stage action alignment network for few-shot action
  recognition.
\newblock In {\em AAAI}, 2022.

\bibitem{lifchitz2019dense}
Yann Lifchitz, Yannis Avrithis, Sylvaine Picard, and Andrei Bursuc.
\newblock Dense classification and implanting for few-shot learning.
\newblock In {\em CVPR}, 2019.

\bibitem{Yang_2022_ECCV}
Yifei~Huang Lijin~Yang and Yoichi Sato.
\newblock Compound prototype matching for few-shot action recognition.
\newblock In {\em ECCV}, 2022.

\bibitem{Muller2007DTW}
Meinard Müller.
\newblock Dynamic time warping.
\newblock {\em Information Retrieval for Music and Motion}, 2007.

\bibitem{trx}
Toby Perrett, Alessandro Masullo, Tilo Burghardt, Majid Mirmehdi, and Dima
  Damen.
\newblock Temporal-relational crosstransformers for few-shot action
  recognition.
\newblock In {\em CVPR}, 2021.

\bibitem{Rusu2019LatentEmbedding}
Andrei~A. Rusu, Dushyant Rao, Jakub Sygnowski, Oriol Vinyals, Razvan Pascanu,
  Simon Osindero, and Raia Hadsell.
\newblock Meta-learning with latent embedding optimization.
\newblock In {\em ICLR}, 2019.

\bibitem{Snell2017Proto}
Jake Snell, Kevin Swersky, and Richard Zemel.
\newblock Prototypical networks for few-shot learning.
\newblock In {\em NeurIPS}, 2017.

\bibitem{ucf101}
Khurram Soomro, Amir~Roshan Zamir, and Mubarak Shah.
\newblock Ucf101: A dataset of 101 human actions classes from videos in the
  wild.
\newblock In {\em CRCV-TR-12-01}, 2012.

\bibitem{su2022temporal}
Bing Su and Ji-Rong Wen.
\newblock Temporal alignment prediction for supervised representation learning
  and few-shot sequence classification.
\newblock In {\em ICLR}, 2022.

\bibitem{Sung2018Compare}
Flood Sung, Yongxin Yang, Li Zhang, Tao Xiang, Philip Torr, and Timothy
  Hospedales.
\newblock Learning to compare: Relation network for few-shot learning.
\newblock In {\em CVPR}, 2018.

\bibitem{thatipelli2021spatio}
Anirudh Thatipelli, Sanath Narayan, Salman Khan, Rao~Muhammad Anwer,
  Fahad~Shahbaz Khan, and Bernard Ghanem.
\newblock Spatio-temporal relation modeling for few-shot action recognition.
\newblock In {\em CVPR}, 2022.

\bibitem{r2plus1d}
Du Tran, Heng Wang, Lorenzo Torresani, Jamie Ray, Yann LeCun, and Manohar
  Paluri.
\newblock A closer look at spatiotemporal convolutions for action recognition.
\newblock In {\em CVPR}, 2018.

\bibitem{vinyals2016matching}
Oriol Vinyals, Charles Blundell, Timothy Lillicrap, koray kavukcuoglu, and Daan
  Wierstra.
\newblock Matching networks for one shot learning.
\newblock In {\em NeurIPS}, 2016.

\bibitem{Wang_2022_CVPR}
Xiang Wang, Shiwei Zhang, Zhiwu Qing, Mingqian Tang, Zhengrong Zuo, Changxin
  Gao, Rong Jin, and Nong Sang.
\newblock Hybrid relation guided set matching for few-shot action recognition.
\newblock In {\em CVPR}, 2022.

\bibitem{wang2019simpleshot}
Yan Wang, Wei-Lun Chao, Kilian~Q Weinberger, and Laurens van~der Maaten.
\newblock Simpleshot: Revisiting nearest-neighbor classification for few-shot
  learning.
\newblock {\em arXiv preprint arXiv:1911.04623}, 2019.

\bibitem{Wu_2022_CVPR}
Jiamin Wu, Tianzhu Zhang, Zhe Zhang, Feng Wu, and Yongdong Zhang.
\newblock Motion-modulated temporal fragment alignment network for few-shot
  action recognition.
\newblock In {\em CVPR}, 2022.

\bibitem{tsl}
Y. Xian, B. Korbar, M. Douze, L. Torresani, B. Schiele, and Z. Akata.
\newblock Generalized few-shot video classification with video retrieval and
  feature generation.
\newblock {\em IEEE TPAMI}, 2021.

\bibitem{Zhang2020FewShotAR}
Hongguang Zhang, Li Zhang, Xiaojuan Qi, Hongdong Li, Philip H.~S. Torr, and
  Piotr Koniusz.
\newblock Few-shot action recognition with permutation-invariant attention.
\newblock In {\em ECCV}, 2020.

\bibitem{zhu2018cmn}
Linchao Zhu and Yi Yang.
\newblock Compound memory networks for few-shot video classification.
\newblock In {\em ECCV}, 2018.

\bibitem{Zhu2021pal}
Xiatian Zhu, Antoine Toisoul, Juan-Manuel Pérez-Rúa, Li Zhang, Brais
  Martinez, and Tao Xiang.
\newblock Few-shot action recognition with prototype-centered attentive
  learning.
\newblock In {\em BMVC}, 2021.

\bibitem{Zhu2021CloserLook}
Zhenxi Zhu, Limin Wang, Sheng Guo, and Gangshan Wu.
\newblock A closer look at few-shot video classification: {A} new baseline and
  benchmark.
\newblock In {\em BMVC}, 2021.

\end{thebibliography}
}
\clearpage
\appendix

\section{Appendix}

In this appendix, we present more formally the matching functions used as baselines for our study in (Section~\ref{sec:supp_matching}), as well as extra experiments that study the impact of different hyper-parameters and present results of different task setups (Section~\ref{sec:supp_ablations}).

\begin{figure*}[t]
\centering 
\begin{tabular}{cc}
    \vspace{-5pt}  
    \begin{subfigure}{0.63\textwidth}
        \includegraphics[width=\textwidth]{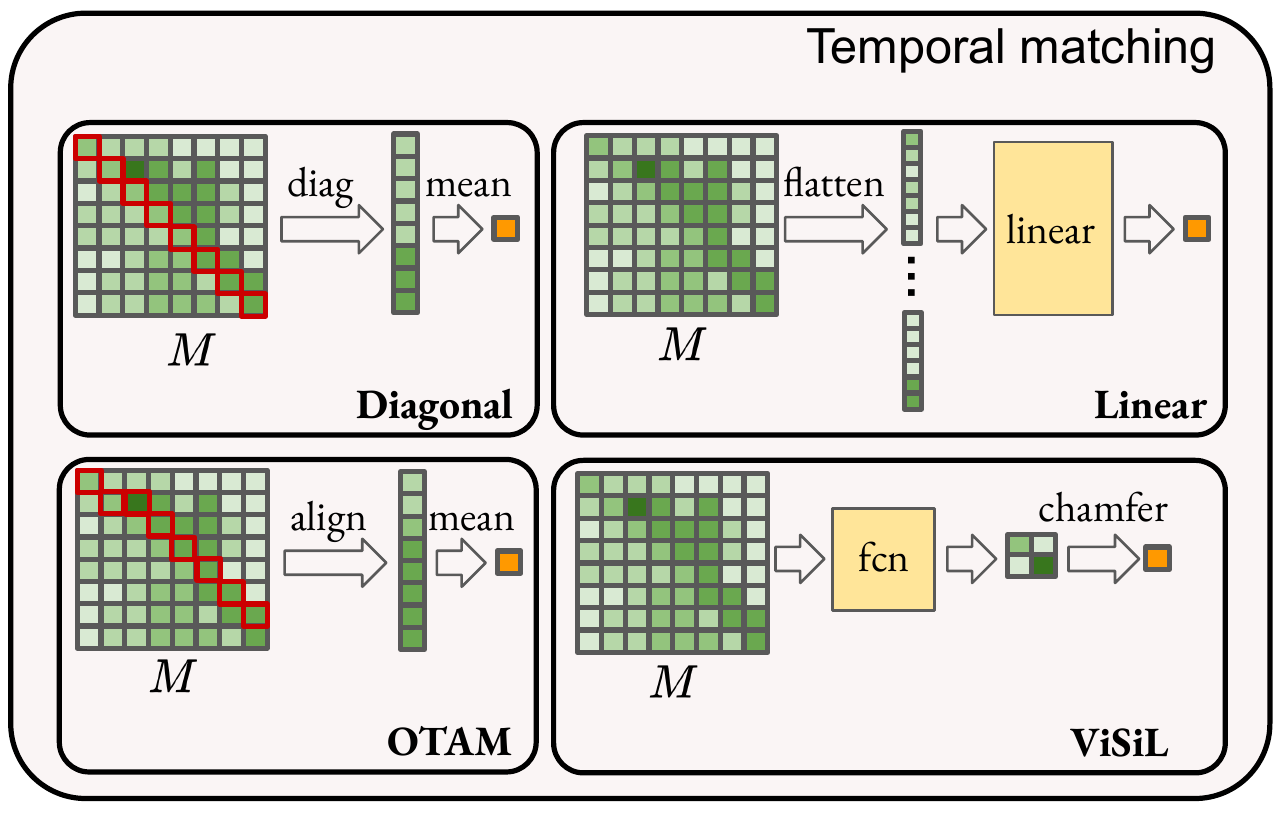}
        \vspace{-15pt}
        \caption{}
        \label{fig:matching_learning}
    \end{subfigure}
    \begin{subfigure}{0.3\textwidth}
        \includegraphics[width=\textwidth]{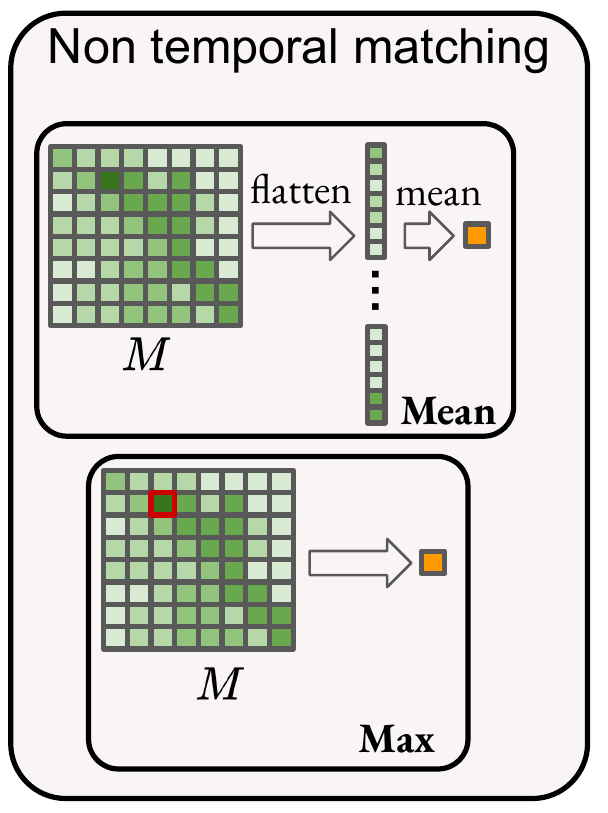}
        \vspace{-15pt}
        \caption{}
        \label{fig:non_temporal_matching_learning}
    \end{subfigure}
\end{tabular}
    \vspace{-2ex}
    \caption{\textbf{Matching functions on the temporal similarity matrix $M$.} We show how each method estimates a scalar video-to-video similarity given the input pairwise similarity matrix. The functions can be classified as a) temporal or b) non-temporal whether they use the temporal position of the features or not.
    \label{fig:matching}
    \vspace{-3ex}
    }
\end{figure*}
\subsection{Baseline matching functions}
\label{sec:supp_matching}

In this section, we describe the different matching functions used as our study's baseline\alert{, and depicted in Figure~\ref{fig:matching}}.
Some matching functions use temporal information by leveraging the absolute or relative position of the pairwise similarities $m_{ij}$. The matching functions that use temporal information are called \emph{temporal} whereas the others are called \emph{non-temporal}.

\subsubsection{Temporal matching functions.}

We provide a list of the temporal matching functions implemented in this study as baselines. Some of them were already introduced in prior work.

\noindent
\myparagraph{Diagonal (Diag)} is used as a baseline in prior work \cite{otam}. It is given by $s(M)= \displaystyle \nicefrac{\sum_{ij} m_{ii}}{n}$. It assumes temporally aligned video pairs.

\noindent
\myparagraph{OTAM}~\cite{otam} uses and extends Dynamic Time Warping~\cite{Muller2007DTW} to find an alignment path on $M$ over which similarities are averaged to produce the video-to-video similarity. A differentiable variant is used for training.

\noindent
\myparagraph{Flatten+FC (Linear)} is a simple baseline we use to learn temporal matching by flattening $M$ and feeding it to a Fully Connected (FC) layer without bias and with a single scalar output. Video-to-video similarity is therefore given by $s(M) = \sum_{ij} w_{ij} m_{ij}$, where $w_{ij}$ are learnable parameters which are $n^2$ in total.

\noindent
\myparagraph{ViSiL}~\cite{visil} is an approach originally introduced for the task of video retrieval. We apply it to few-shot action recognition for the first time. A small Fully Convolutional Network (FCN) is applied on $M$. Its output is a filtered temporal similarity matrix, and the Chamfer similarity is applied on it. The filtering is performed according to the small temporal context captured by the small receptive field of this network.

\subsubsection{Non-temporal matching functions.} 
We provide a list of the non-temporal matching functions that were implemented in this study as baselines. 

\noindent
\myparagraph{Mean} is used as a baseline in prior work \cite{otam}. It is given by $s(M)= \displaystyle \nicefrac{\sum_{ij} m_{ij}}{n^2}$. It supposes all the clip pairs should contribute equally to the similarity score.

\noindent
\myparagraph{Max} is used as a baseline in prior work \cite{otam}. It is given by $s(M)= \max_{ij} m_{ij}$.
It supposes that selecting the best matching clip pair is enough to recognize the action.

\subsection{Additional ablations and impact of hyper-parameters}
\label{sec:supp_ablations}
In this section, we present additional ablations to evaluate the impact of the feature projection head, the ordering of the tuples, and the number of examples per class used in the support set. We also report the impact of using the different variants for the \kinetics and \ucf datasets.

\begin{table}[t]
\caption{\textbf{Impact of learning a feature projection on performance of matching-based methods}. $^\dagger$ denotes hand-crafted matching methods, \ie no training is performed for the cases where a feature projection is not learned.
\label{tab:feature_embeddings_dataloader}
}
\vspace{1ex}
\centering
\vspace{2ex}

\setlength\extrarowheight{3pt}
    \resizebox{\linewidth}{!}{
    \setlength{\tabcolsep}{12pt}
    \begin{tabular}{ l c cc cc} 
    \toprule
         \multirow{2}{*}{Method}  & Learned       & \multicolumn{2}{c}{\ssvtwo} & \multicolumn{2}{c}{\kinetics}  \\
                                  & Proj.       &  \oneshot & \fiveshot & \oneshot & \fiveshot \\
         \midrule
         Max$^\dagger$ &    & 63.40 & 75.80 & 94.90 & 97.50  \\ 
         Max &  \checkmark & 64.97 & 78.97 &  95.27 & 98.30   \\
         & & \diffup{1.57} & \diffup{3.17} & \diffup{0.37} & \diffup{0.80} \\
          \midrule
         Chamfer++ (l=3)$^\dagger$ &    & 64.50 & 79.50 & 94.10 & 98.10  \\ 
         Chamfer++ (l=3) &  \checkmark  & 67.83 & 81.60 &  96.10 & 98.30   \\
         & & \diffup{3.33} & \diffup{2.10} & \diffup{2.00} & \diffup{0.20} \\
          \midrule
         Mean$^\dagger$ &    & 61.40 & 75.40 & 92.50 & 97.50  \\ 
         Mean &  \checkmark & 65.77 & 79.13 &  95.53 & 98.10   \\
         & & \diffup{4.37} & \diffup{3.73} & \diffup{3.03} & \diffup{0.60} \\
          \midrule
         OTAM~\cite{otam}$^\dagger$ &  & 63.70  & 76.40 & 93.70 & 97.90 \\ 
         OTAM~\cite{otam} & \checkmark & 67.10  & 80.23 & 95.93 & 98.37   \\ 
         & & \diffup{3.40} & \diffup{3.83} & \diffup{2.23} & \diffup{0.47} \\
    \bottomrule
    \end{tabular}
    }

\vspace{1ex}
\end{table}

\begin{table}[t]
\vspace{1ex}
\caption{\textbf{Impact of the dimension size of the feature projection head} for Chamfer++ using ordered-tuples and $l=3$.
\label{tab:feature_projection_dimension}
}
\centering

\vspace{2ex}

\resizebox{\linewidth}{!}
{
\small
\setlength\extrarowheight{-1pt}
\setlength{\tabcolsep}{12pt}
\begin{tabular}{ l c cc cc} 
\toprule
    \multirow{2}{*}{Method}  & $D$ &  \multicolumn{2}{c}{\ssvtwo} & \multicolumn{2}{c}{\kinetics}  \\
    & &  \oneshot & \fiveshot & \oneshot & \fiveshot  \\
     \midrule
     Chamfer++ (l=3)  &  & 64.5 \stddev{0.0}& 79.5 \stddev{0.0} & 94.1 \stddev{0.0}& 98.1 \stddev{0.0} \\
     Chamfer++ (l=3)  &  512 & 67.3 \stddev{0.2} & 81.2 \stddev{0.0} & 95.9 \stddev{0.1} &  98.2 \stddev{0.1}  \\
     Chamfer++ (l=3)  &  1024 & 67.8 \stddev{0.3} & 81.3  \stddev{0.1} & 96.1 \stddev{0.2} & 98.3 \stddev{0.0}  \\
     Chamfer++ (l=3)  &  1152 &  67.8 \stddev{0.2} & 81.6 \stddev{0.1} & 96.1 \stddev{0.1}& 98.3 \stddev{0.0}  \\
     Chamfer++ (l=3)  &  2048 & 67.9  \stddev{0.0} & 81.8 \stddev{0.2} & 96.1 \stddev{0.1} &  98.3 \stddev{0.0}  \\
\bottomrule
\end{tabular}
    }

\vspace{1ex}
\end{table}

\subsubsection{The impact of the projection layer} for matching methods is validated in Table~\ref{tab:feature_embeddings_dataloader}.
The performance is consistently improved on all setups and methods by including and learning a projection layer.
Although the backbone is trained with TSL on the same meta-train set, the projection layer allows features and values in the temporal similarity matrix to better align with each matching process.

\subsubsection{Projection head dimension.}
To match the work from \cite{trx}, we set the projection dimension to $D=1152$. This section evaluates the effect of using different values for $D$. The results are reported in Table~\ref{tab:feature_projection_dimension}. A minimum value of $D=1024$ seems enough and could be used for future experiments.

\subsubsection{Impact of using different variants}

\label{sec:supplementary_ablation}
\begin{table}[t]
\caption{Chamfer++ variants on the three datasets: \ucf, \kinetics and \ssvtwo. 
\label{tab:chamfer_ablation_all}}
\centering

\resizebox{1.0\linewidth}{!}
{
\small
\setlength\extrarowheight{-1pt}
\setlength{\tabcolsep}{12pt}
\begin{tabular}{@{\msp} l@{\msp} l@{\msp}l@{\msp} l@{\msp}l@{\msp} l@{\msp}l@{\msp} l@{\msp}l@{\msp} l@{\msp}} 
\toprule
     \multirow{2}{*}{Method}    & \multirow{2}{*}{$l$}    & \multirow{2}{*}{Joint}    &
     \multirow{2}{*}{Tupl.} & \multicolumn{2}{c}{\ssvtwo} & \multicolumn{2}{c}{\kinetics} & \multicolumn{2}{c}{\ucf}  
     \\
      &  & &  & \oneshot & \fiveshot & \oneshot & \fiveshot  & \oneshot & \fiveshot \\
     \midrule
     Chamfer-Q  &  1 & & &  65.7 \stddev{0.1} & 79.7 \stddev{0.1} & 95.5 \stddev{0.1} & 98.1 \stddev{0.1} & 97.8 \stddev{0.1} & 99.0 \stddev{0.1}   \\
    Chamfer-S  &  1 & &  & 65.3 \stddev{0.1} & 79.1 \stddev{0.2} & 95.4 \stddev{0.0} & 98.2 \stddev{0.1}  & 97.7 \stddev{0.2}& 98.9 \stddev{0.1}   \\
     Chamfer  &  1 &    &  & 66.9 \stddev{0.1} & 80.0 \stddev{0.2}  & 96.0 \stddev{0.1} & 98.3 \stddev{0.1} & 97.9 \stddev{0.2} & 99.0 \stddev{0.1}   \\
     Chamfer+  &  1 &  \checkmark &   & 66.9 \stddev{0.1} & 80.7 \stddev{0.2}  & 96.0 \stddev{0.1} &  98.2 \stddev{0.1}  & 97.9 \stddev{0.2} &  99.2 \stddev{0.1} \\

     Chamfer++  &  2 &  \checkmark  & all & 67.1 \stddev{0.1} & 80.8 \stddev{0.2}  & 96.1 \stddev{0.1} & 98.3 \stddev{0.1} & 97.8 \stddev{0.1} & 99.2 \stddev{0.1}  \\
     Chamfer++  &  2 &  \checkmark  & ord. & 67.7 \stddev{0.1} & 81.4 \stddev{0.2}  & 96.1 \stddev{0.1} & 98.3 \stddev{0.1} & 97.9 \stddev{0.1} & 99.2 \stddev{0.1}  \\
     Chamfer++  &  3 &  \checkmark & all & 67.0 \stddev{0.3} & 80.8 \stddev{0.1}  & 96.2 \stddev{0.2}& 98.4 \stddev{0.0} & 97.8 \stddev{0.1} & 99.2 \stddev{0.1}  \\
     Chamfer++  &  3 &  \checkmark & ord. & 67.8 \stddev{0.2} & 81.6 \stddev{0.1}  & 96.1 \stddev{0.1} & 98.3 \stddev{0.0} & 97.7 \stddev{0.0} & 99.3 \stddev{0.0}  \\
    
\bottomrule
\end{tabular}
    }

\end{table}

We report the accuracy for the different variants of Chamfer++ for the \kinetics and \ucf datasets in Table~\ref{tab:chamfer_ablation_all}.  
As for \ssvtwo, both variants improve performances compared to the vanilla approach.

\subsubsection{Impact of ordering the clip-tuples}

In this section, we evaluate the impact of using ordered clip feature tuples $\vt^{l}$ versus using all the clip feature tuples $\vt_{all}^{l}$.
The comparison between ordered tuples and all tuples is presented in Table~\ref{tab:tuple_order}. On the \ssvtwo dataset, using ordered clip feature tuples boosts the accuracy. On the \kinetics and the \ucf datasets, using ordered tuples doesn't provide a boost and can even slightly harm the performance. 
Since the number of tuples is significantly lower when they are in order, using the ordered clip feature tuples is preferable.

\begin{table}[t]
\vspace{1ex}
\caption{\textbf{Impact of using ordered clip feature tuples vs all the features} for different values of $l$.
\label{tab:tuple_order}
}
\centering
\vspace{2ex}

\setlength\extrarowheight{3pt}
    \resizebox{\linewidth}{!}{
    \setlength{\tabcolsep}{12pt}
    \begin{tabular}{ l c c cc cc cc} 
    \toprule
         \multirow{2}{*}{Method}  & \multirow{2}{*}{$l$} & Ordered &\multicolumn{2}{c}{\ssvtwo} & \multicolumn{2}{c}{\kinetics} & \multicolumn{2}{c}{\ucf}  \\
          & & Tuples&  \oneshot & \fiveshot & \oneshot & \fiveshot & \oneshot & \fiveshot \\
         \midrule
         Chamfer++ & 2 &  & 67.10 & 80.80 & 96.10 & 98.33 & 97.80 & 99.23   \\ 
         Chamfer++ & 2 & \checkmark & 67.73 & 81.40 & 96.10 & 98.33 & 97.87 & 99.23   \\ 
         & & & \diffup{0.63} & \diffup{0.80} & \diffup{0.00} & \diffup{0.00} & \diffup{0.07} & \diffup{0.00}  \\
         \midrule
         Chamfer++ & 3 &  & 67.03 & 80.83 & 96.17 & 98.37 & 97.77 & 99.23   \\ 
         Chamfer++ & 3 & \checkmark & 67.83 & 81.60 & 96.10 & 98.30 & 97.73 & 99.27   \\ 
         & & & \diffup{0.80} & \diffup{0.77} &  \diffdown{0.07} & \diffdown{0.07} & \diffdown{0.04} & \diffup{0.04}  \\
        
    \bottomrule
    \end{tabular}
    }

\vspace{1ex}
\end{table}

\subsubsection{Impact of the number of examples per class}

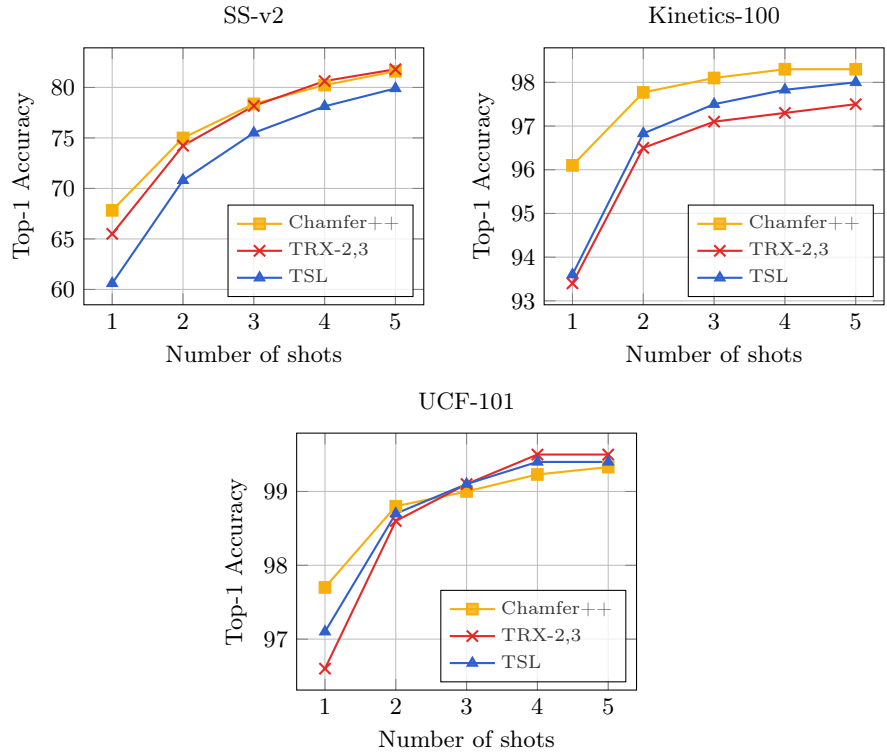
\begin{figure*}
\centering
\begin{tabular}{cc}
\definecolor{C1}{RGB}{226, 43, 41} 
\definecolor{C2}{RGB}{47, 96, 206} 
\definecolor{C3}{RGB}{246, 175, 11} 

\begin{tikzpicture}
  \begin{axis}[
    width=0.5\linewidth,
    height=5cm, 
    xtick = {0,1,2,3,4,5},
  	legend cell align={left},
	legend pos=south east,
    legend style={font=\scriptsize}, 
    grid=both,
    xlabel={\small Number of shots},
    ylabel={\small Top-1 Accuracy},
    title={\ssvtwo}
  ]
    \addplot [thick, color=C3, mark=square*,  mark size=2] coordinates {(1, 67.83) (2, 75) (3, 78.37) (4, 80.23) (5, 81.60)}; 
	\addlegendentry{Chamfer++} 
	
	\addplot [thick, color=C1,  mark=x,  mark size=3] coordinates {(1, 65.5) (2, 74.23) (3, 78.17) (4, 80.63) (5, 81.8)}; 
	\addlegendentry{TRX-{2,3}} 
	  
    \addplot [thick, color=C2, mark=triangle*,  mark size=2] coordinates {(1, 60.6) (2, 70.8) (3, 75.5) (4, 78.13) (5, 79.9)};  
	\addlegendentry{TSL}
	
  \end{axis}
\end{tikzpicture}
\definecolor{C1}{RGB}{226, 43, 41} 
\definecolor{C2}{RGB}{47, 96, 206} 
\definecolor{C3}{RGB}{246, 175, 11} 

\begin{tikzpicture}
  \begin{axis}[
    width=0.5\linewidth,
    height=5cm, 
    xtick = {0,1,2,3,4,5},
  	legend cell align={left},
	legend pos=south east,
    legend style={font=\scriptsize}, 
    grid=both,
    xlabel={\small Number of shots},
    ylabel={\small Top-1 Accuracy},
    title={\kinetics}
  ]
    \addplot [thick, color=C3, mark=square*,  mark size=2] coordinates {(1, 96.10) (2, 97.77) (3, 98.1) (4, 98.3) (5, 98.30)}; 
	\addlegendentry{Chamfer++} 
	
	\addplot [thick, color=C1,  mark=x,  mark size=3] coordinates {(1, 93.4) (2, 96.5) (3, 97.1) (4, 97.3) (5, 97.5)}; 
	\addlegendentry{TRX-{2,3}}]
    
    \addplot [thick, color=C2, mark=triangle*,  mark size=2] coordinates {(1, 93.6) (2, 96.83) (3, 97.5) (4, 97.83) (5, 98)};  
	\addlegendentry{TSL} 
 
  \end{axis}
\end{tikzpicture}
\end{tabular}
\definecolor{C1}{RGB}{226, 43, 41} 
\definecolor{C2}{RGB}{47, 96, 206} 
\definecolor{C3}{RGB}{246, 175, 11} 

\begin{tikzpicture}
  \begin{axis}[
    width=0.5\linewidth,
    height=5cm, 
    xtick = {0,1,2,3,4,5},
  	legend cell align={left},
	legend pos=south east,
    legend style={font=\scriptsize}, 
    grid=both,
    xlabel={\small Number of shots},
    ylabel={\small Top-1 Accuracy},
    title={\ucf}
  ]
    \addplot [thick, color=C3, mark=square*,  mark size=2] coordinates {(1, 97.7) (2, 98.8) (3, 99) (4, 99.23) (5, 99.33)}; 
	\addlegendentry{Chamfer++}  
	
	\addplot [thick, color=C1,  mark=x,  mark size=3] coordinates {(1, 96.6) (2, 98.6) (3, 99.1) (4, 99.5) (5, 99.5)};   
	\addlegendentry{TRX-{2,3}}]
    
    \addplot [thick, color=C2, mark=triangle*,  mark size=2] coordinates {(1, 97.1)(2, 98.7) (3, 99.1) (4, 99.4) (5, 99.4)};  
	\addlegendentry{TSL}  
	
  \end{axis}
\end{tikzpicture}
\vspace{-0.5ex}
\caption{\textbf{Evolution of the accuracy with the number of examples per class} in the support set for the three datasets.
\label{fig:multi-shots}
}
\end{figure*}

The impact of $k$ is shown in Figure~\ref{fig:multi-shots} by measuring performance for an increasing number of support examples per class while keeping the number of classes fixed, $C_{t}=C_{f}=5$. We observe that TSL and TRX have inferior performances for the low-shot regime, while their performance increases faster with the number of shots. In the low-regime, Chamfer-QS++ outperforms the other methods and still keeps some benefits while the number of shots increases.

\end{document}